\documentclass[10pt,twocolumn,letterpaper]{article}

\usepackage[pagenumbers]{cvpr} 

\usepackage[dvipsnames]{xcolor}
\definecolor{airforceblue}{rgb}{0.36, 0.54, 0.66}

\newcommand{\comm}[1]{\iffalse #1 \fi}
\usepackage{colortbl}
\usepackage{soul}
\usepackage{pifont}
\usepackage{lipsum}
\usepackage{hhline}
\usepackage{booktabs}
\usepackage{subcaption}
\usepackage{multirow}
\usepackage{multicol}
\usepackage{xspace}
\usepackage{float}

\newcommand{\xmark}{\ding{55}}%
\definecolor{Gray}{gray}{0.9}
\newcommand{\model}{\textsc{DiffAnt}\xspace}
\newcommand{\myparagraph}[1]{\vspace{3.5pt}\noindent\textbf{#1}}

\definecolor{cvprblue}{rgb}{0.21,0.49,0.74}
\usepackage[pagebackref,breaklinks,colorlinks,citecolor=cvprblue]{hyperref}

\def\paperID{11359} 
\def\confName{CVPR}
\def\confYear{2024}

\title{\model: Diffusion Models for Action Anticipation}

\author{
Zeyun Zhong$^{1,2}$ \quad
Chengzhi Wu$^1$\quad 
Manuel Martin$^2$\quad 
Michael Voit$^2$\quad
Juergen Gall$^3$\quad 
Jürgen Beyerer$^{1,2}$\\
$^1$Karlsruhe Institute of Technology \qquad $^2$Fraunhofer IOSB \qquad $^3$University of Bonn \\
{\tt\small \{firstname.lastname\}@\{kit.edu, iosb.fraunhofer.de\}}
\quad
{\tt\small gall@iai.uni-bonn.de}
}

\begin{document}
\maketitle
\begin{abstract}

Anticipating future actions is inherently uncertain. Given an observed video segment containing ongoing actions, multiple subsequent actions can plausibly follow. This uncertainty becomes even larger when predicting far into the future. However, the majority of existing action anticipation models adhere to a deterministic approach, neglecting to account for future uncertainties. In this work, we rethink action anticipation from a generative view, employing diffusion models to capture different possible future actions.
In this framework, future actions are iteratively generated from standard Gaussian noise in the latent space, conditioned on the observed video, and subsequently transitioned into the action space.
Extensive experiments on four benchmark datasets, i.e., Breakfast, 50Salads, EpicKitchens, and EGTEA Gaze+, are performed and the proposed method achieves superior or comparable results to state-of-the-art methods, showing the effectiveness of a generative approach for action anticipation.
Our code and trained models will be published on GitHub.
\end{abstract}    
\section{Introduction}

\begin{figure}[t]
    \centering
    \includegraphics[width=\linewidth]{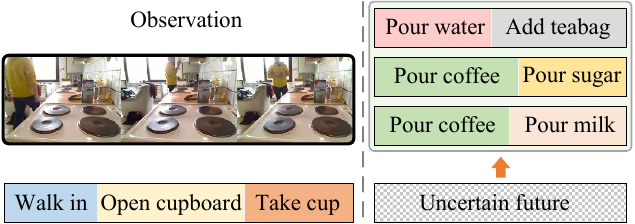}
    \caption{The inherent uncertainty in predicting future actions presents a complex scenario where, given a single observation, multiple feasible\comm{viable} future action sequences may emerge. Example frames are from the Breakfast~\cite{kuehneLanguageActionsRecovering2014} dataset.}
    \label{fig:fig1}
\end{figure}

In contexts such as human-machine cooperation and robotic assistance, the anticipation of potential future daily-living actions is vital. For instances where timely assistance is needed or proactive dialogues are essential, the capability to accurately predict actions, even in the absence of direct observation, is of utmost importance. Yet, the ever-changing landscape of daily activities introduces a natural unpredictability to future actions, as depicted in Fig.~\ref{fig:fig1}. This inherent uncertainty becomes even larger if we are going to predict far into the future, which poses significant challenges to the precise anticipation of future actions. Consequently, modeling the underlying uncertainty may be beneficial, allowing to capture different possible futures.

Addressing the intricate challenges and uncertainties of action anticipation calls for a departure from conventional methods. Traditional approaches~\cite{farhaWhenWillYou2018,keTimeConditionedActionAnticipation2019,senerTemporalAggregateRepresentations2020,gongFutureTransformerLongterm2022} predominantly employ a discriminative and deterministic stance, framing action prediction as a classification task with deterministic future outcomes. This approach, however, can inadvertently diminish the visibility of plausible alternative actions~\cite{zhao2020diverse}. Although probabilistic models~\cite{abufarhaUncertaintyAwareAnticipationActivities2019,mehrasaVariationalAutoEncoderModel2019,zhao2020diverse}, including
generative models like GANs~\cite{zhao2020diverse} and VAEs~\cite{mehrasaVariationalAutoEncoderModel2019}, have been introduced to embrace the non-deterministic essence of future anticipations, they rely on action labels from observed frames, 
thereby constraining their immediate applicability. As such, a model that acknowledges inherent uncertainties and supports direct deployment is of paramount importance.

Within this context, we rethink action anticipation from a generative view, introducing an end-to-end probabilistic generative model, \model. Leveraging diffusion models~\cite{ho2020ddpm,song2021ddim}, which have emerged as a paradigm with significant promise across various domains, our approach iteratively generates future actions in the latent space from the standard Gaussian noise. To adapt the diffusion models for action anticipation, we extend the standard models by introducing a future action embedding function and an action predictor. Moreover, to enhance the precision of our predictions, we incorporate visual observations as conditional information and employ an encoder-decoder structure~\cite{vaswaniAttentionAllYou2017,carion2020end,gongFutureTransformerLongterm2022} for seamless integration. We evaluate \model on four standard benchmarks for long-term action anticipation, achieving state-of-the-art results on Breakfast, 50Salads, and EGTEA Gaze+, and comparable results on EpicKitchens.
In summary, our main contributions are:

\begin{itemize}
    \item A probabilistic generative approach, \model, for long-term action anticipation, employing diffusion models and leveraging their stochastic and continuous nature to iteratively refine predictions and navigate the intrinsic uncertainties that exist in predicting future actions.
    \item We conduct extensive experiments on four widely used benchmark datasets, \ie, Breakfast, 50Salads, EpicKitchens, and EGTEA Gaze+. Our results consistently demonstrate superior or comparable performance to state-of-the-art deterministic methods.
    \item We present an in-depth analysis of our approach, demonstrating that our model surpasses the current state-of-the-art probabilistic method.
\end{itemize}
\section{Related Work}
\noindent\textbf{Action Anticipation} aims to predict future actions given a video clip of the past and present. Many approaches initially investigated different forms of action and activity anticipation from third person video~\cite{gaoREDReinforcedEncoderDecoder2017,farhaWhenWillYou2018,keTimeConditionedActionAnticipation2019,gongFutureTransformerLongterm2022}. Recently, along with development of multiple challenge benchmarks~\cite{damen2018scaling,damen2020epic,li2018eye,graumanEgo4DWorld0002022}, the first-person (egocentric) vision has also gained popularity.
To accurately predict future actions, the summarization of temporal progression of past actions is essential. To model the past action progression, earlier methods mainly used RNN~\cite{farhaWhenWillYou2018,furnariWhatWouldYou2019} or TCN~\cite{keTimeConditionedActionAnticipation2019}-based architectures, which have been shown to be inferior to the recent Transformer-based approaches~\cite{girdharAnticipativeVideoTransformer2021,zhong2023afft,gongFutureTransformerLongterm2022,nawhal2022rethinking}.
Based on the predicted time horizon, action anticipation approaches can be broadly grouped into two categories~\cite{zhong2023survey}: short-term anticipation approaches~\cite{damen2018scaling,damen2020epic} and long-term anticipation approaches~\cite{farhaWhenWillYou2018,graumanEgo4DWorld0002022}. While short-term approaches predict actions a few seconds into the future, long-term approaches aim to predict a sequence of future actions (with their durations) up to several minutes into the future.

\myparagraph{Long-term Anticipation.} 
An initial work~\cite{farhaWhenWillYou2018} introduced two distinct models for long-term action anticipation. While the RNN model performed in a recursive manner, the CNN model outputs a sequence of future actions in the form of a matrix in one single step. Ke~\etal~\cite{keTimeConditionedActionAnticipation2019} developed a model targeting the prediction of a specific future action, bypassing the need for anticipating intermediate actions to avoid error accumulations. Farha~\etal~\cite{abu2020long} introduced a cycle consistency module to predict past activities using the projected future, demonstrating improved outcomes in comparison to its counterpart lacking the consistency module. Sener~\etal~\cite{senerTemporalAggregateRepresentations2020} suggested a multi-scale temporal aggregation model that aggregates past visual features in condensed vectors and then iteratively predicts future actions using an LSTM. Recently, Transformer-based approaches~\cite{gongFutureTransformerLongterm2022,nawhal2022rethinking} have been also employed for long-term anticipation.
Different from these approaches, our method is non-deterministic and is capable to take the uncertain future into account.

\myparagraph{Uncertainty-aware Anticipation.}
To address the inherent uncertainty involved in forecasting future actions, various researches have proposed non-deterministic approaches, including generating an array of potential outcomes with multiple rules~\cite{vondrick2016anticipating,piergiovanni2020adversarial} and by sampling from the learned distribution~\cite{abufarhaUncertaintyAwareAnticipationActivities2019,mehrasaVariationalAutoEncoderModel2019,zhao2020diverse,mascaro2023intention}. Vondrick~\etal~\cite{vondrick2016anticipating} proposed training a mixture of networks, each aiming to predict one potential future.
Piergiovanni~\etal~\cite{piergiovanni2020adversarial} proposed a differentiable grammar model and applied adversarial techniques to enable efficient learning and avoid enumerating all possible rules. 
Farha and Gall~\cite{abufarhaUncertaintyAwareAnticipationActivities2019} 
anticipated all subsequent actions and durations stochastically, employing an action model and a time model trained to predict the probability distribution of future action labels and durations. In the literature, probabilistic generative models like GANs~\cite{zhao2020diverse} and VAEs~\cite{mehrasaVariationalAutoEncoderModel2019} have been employed to capture the uncertain aspects of future predictions. Nonetheless, these methods typically require the action labels of observed frames as inputs, which are obtained either from ground truth labels or inferred through an action segmentation model~\cite{ding2022surveyactionsegmentation}.
Drawing inspiration from these works, we approach the anticipation task from a generative perspective, employing diffusion models to generate diverse and plausible future actions, while leveraging the rich visual features.

\myparagraph{Diffusion Models.} Initially unified with score-based models~\cite{score1,score2,score3}, diffusion models~\cite{Therm,ho2020ddpm,song2021ddim} are renowned for their stable training processes, which do not rely on adversarial mechanism for generative learning. These models have demonstrated remarkable achievements across various domains, including image generation~\cite{dhariwal2021beatsGAN,rombach2022latentdiffusion,wang2023learning}, natural language generation~\cite{yu2022latent,li2022diffusionlm,gong2023diffuseq}, text-to-image synthesis~\cite{gu2022vector,kim2021diffusionclip}, and audio generation~\cite{leng2022binauralgrad,lam2022bddm}.
Recent advances have extended the application of diffusion models to image and video comprehension tasks within computer vision, such as object detection~\cite{DiffusionDet}, image segmentation~\cite{2021_ICLR_Baranchuk,SegDiff}, video forecasting and infilling~\cite{yang2022diffusion,hoppe2022diffusion,voleti2022masked}, and action segmentation~\cite{liu2023diffact}.
In this work, we leverage the iterative refinement capabilities of diffusion models for future prediction. To the best of our knowledge, this work is the first one employing diffusion models for action anticipation.
\newcommand{\mathbfz}{\mathbf{z}}
\newcommand{\boldmu}{\boldsymbol{\mu}_\theta(\mathbfz_{s},s)}
\newcommand{\fzpred}{f_{\theta}(\mathbfz_{s},s)}

\section{Diffusion Action Anticipation}
\begin{figure}[t]
    \centering
    \includegraphics[width=\linewidth]{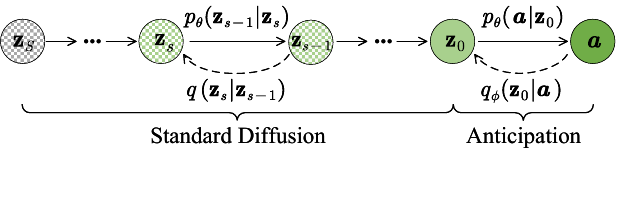}\vspace{-.85cm}
    \caption{Concept of \model. Standard diffusion models reconstruct original data $\mathbf{z}_0$ from pure noise $\mathbf{z}_S$. Our approach integrates discrete future actions $\boldsymbol{a}$ by introducing an embedding function and a predictor to facilitate the conversion between continuous $\mathbf{z}_0$ and discrete $\boldsymbol{a}$.}
    \label{fig:concept}
\end{figure}

To address the inherently non-deterministic nature of future action anticipation, we present \model, a novel diffusion model for action anticipation. The model integrates an action embedding function and an action predictor into the original diffusion framework, enabling the incorporation of discrete future actions within both the forward and reverse diffusion processes (see Fig.~\ref{fig:concept}). To refine the predictive capabilities of \model, an encoder-decoder architecture~\cite{vaswaniAttentionAllYou2017,carion2020end,gongFutureTransformerLongterm2022} is utilized to assimilate visual observations as conditional information.
We acquaint the reader with the problem statement and the background of diffusion models in Section~\ref{sec:preliminary}. Subsequently, the inference and training processes of the proposed method are introduced in Section~\ref{sec:inference} and \ref{sec:training}, respectively.

\subsection{Preliminaries}
\label{sec:preliminary}

\noindent\textbf{Problem Statement.}
Given the past observation containing observed video features $F \in \mathbb{R}^{L \times K}$ with $K$ dimensions for $L$ frames, the long-term anticipation task aims to predict following $N$ future actions $\boldsymbol{a}$ within a video, consisting of action classes $\boldsymbol{a}^c \in \mathbb{R}^{N\times C}$ and durations $\boldsymbol{a}^t \in \mathbb{R}^{N\times 1}$, where $C$ is the total number of action classes.

\myparagraph{Diffusion Models} aim to approximate the data distribution $q(\mathbfz_0)$ with a model distribution $p_{\theta}(\mathbfz_0)$~\cite{ho2020ddpm,song2021ddim}. A diffusion model typically contains forward and reverse processes.
The forward process or diffusion process corrupts the real data $\mathbfz_0 \sim q(\mathbfz_0)$ into a series of noisy data $\mathbfz_1, \mathbfz_2, \dots$ and finally into a standard Gaussian noise $\mathbfz_S \sim \mathcal{N}(\mathbf{0}, \boldsymbol{I})$. For each forward step $s \in [1, 2, \dots, S]$, the perturbation is controlled by $q(\mathbf{z}_{s} \vert \mathbf{z}_{s-1}) = \mathcal{N}(\mathbf{z}_{s};\sqrt{1-\beta_s}\mathbf{z}_{s-1}, {\beta}_s \boldsymbol{I})$, with predefined $\beta_s \in (0,1)$ as different variance scales. By denoting $\alpha_s = 1 - \beta_s$ and $\bar{\alpha}_s = \prod_{i=1}^{s} \alpha_i$, we can directly obtain $\mathbfz_s$ from $\mathbfz_0$ in a closed form without recursion:
$q(\mathbfz_s|\mathbfz_{0}) = \mathcal{N}(\mathbfz_s;\sqrt{\bar{\alpha}_s}\mathbfz_{0},(1-\bar{\alpha}_s) \boldsymbol{I}),$
which can be further simplified using the reparameterization trick~\cite{kingma2013auto}.

The reverse process or denoising process tries to gradually remove the noise from $\mathbfz_{S} \sim \mathcal{N}(\boldsymbol{0}, \boldsymbol{I})$ to reconstruct the original data $\mathbf{z}_0$. Each reverse step is defined as $
p_\theta(\mathbfz_{s-1}|\mathbfz_{s}) = \mathcal{N}(\mathbfz_{s-1};\boldsymbol{\mu}_\theta(\mathbfz_{s},s), \sigma_s^2 \boldsymbol{I})$,
where $\sigma_s^2$ is controlled by $\beta_s$, and $\boldmu$ is a predicted mean parameterized by a step-dependent diffusion model $f_{\theta}(\mathbfz_s,s)$. Several different ways~\cite{luo2022understanding} are possible to parameterize $p_\theta$, including the prediction of mean $\boldmu$, the prediction of the noise $\epsilon$, and the prediction of $\mathbfz_0$. We choose to predict $\mathbfz_0$, as suggested in several related works~\cite{li2022diffusionlm,gong2023diffuseq,liu2023diffact}.

\begin{figure}[t]
    \centering
    \begin{subfigure}[b]{\linewidth}
    \centering
        \includegraphics[angle=270,origin=c,width=0.75\linewidth]{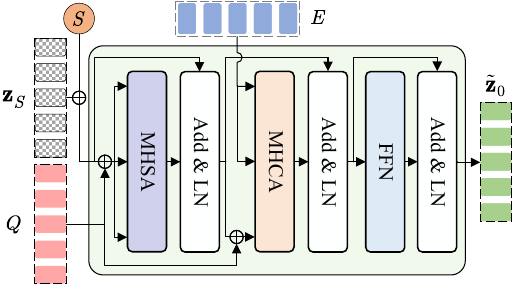}\vspace{-1.5cm}
        \subcaption{Details of a decoder layer.}
        \label{fig:decoder}
    \end{subfigure}\vspace{0.2cm}
    \begin{subfigure}[b]{\linewidth}
    \centering
    \includegraphics[width=\linewidth]{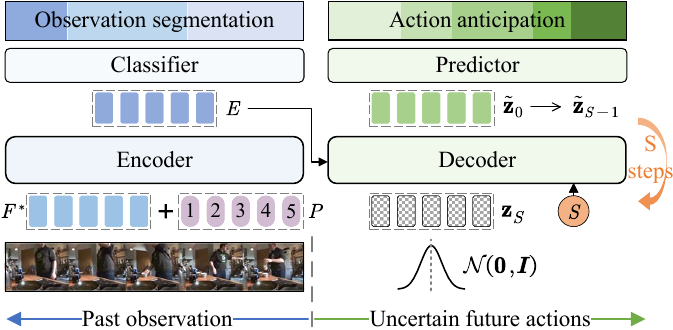}
    \subcaption{Inference pipeline.}    \label{fig:inference_pipeline}
    \end{subfigure}\vspace{-0.1cm}
    \caption{Inference pipeline of \model. (a) The decoder uses action queries $Q$ and step information to refine the noisy futures $\mathbf{z}_S$, conditioned on encoded past observations $E$ via cross-attention. (b) Future action embeddings $\mathbf{z}_S$ are drawn from the standard Gaussian distribution and iteratively denoised to produce the final $\Tilde{\mathbf{z}}_0$, which is then decoded into future action labels and durations.}
    \label{fig:inference}
\end{figure}

\subsection{Overall Inference Pipeline}
\label{sec:inference}
To better harness observed video features $F$ as conditional information for controlled future action generation, we propose an encoder-decoder structure depicted in Fig.~\ref{fig:inference_pipeline}.

\myparagraph{Encoder.}
The input features $F$, typically extracted per short clip by pre-trained models, are first processed through an encoder to incorporate long-range temporal context, making them task-oriented. Although our framework permits flexibility in the choice of the encoder, we opt for the standard transformer model used in DETR~\cite{carion2020end} and FUTR~\cite{gongFutureTransformerLongterm2022}.
We begin by passing the video features $F$ through a linear layer, which adjusts them to a suitable hidden dimension $D$, yielding transformed features $F^*$. These, combined with sinusoidal positional encodings $P \in \mathbb{R}^{L\times D}$, enable the encoder to generate refined representations $E \in \mathbb{R}^{L\times D}$. Such representations are conducive to action anticipation and can be mapped to the action space using a linear classifier, facilitating observation segmentation.

\myparagraph{Decoder.}
Our decoder, as depicted in Fig.~\ref{fig:decoder}, follows the query-based paradigm~\cite{carion2020end,zhu2020deformabledetr}, originally proposed to eliminate handcrafted components in object detection and recently applied in long-term action anticipation~\cite{gongFutureTransformerLongterm2022,nawhal2022rethinking}. It processes noisy action embeddings $\mathbf{z}_S \in \mathbb{R}^{M\times D'}$ and action queries $Q \in \mathbb{R}^{M\times D'}$, generating refined futures $\Tilde{\mathbf{z}}_0$ in parallel, under the guidance of encoded past observations $E$ via cross-attention.
The action queries $Q$ consist of $M$ learnable tokens of dimension $\mathbb{R}^{1\times D'}$, with their temporal order aligned with the sequence of future actions, \ie, each query directly corresponds to a respective future action~\cite{gongFutureTransformerLongterm2022}.

\myparagraph{Predictor.}
The predictor decodes the refined future embeddings $\Tilde{\mathbf{z}}_0$ into action labels $\Tilde{\mathbf{a}}^c \in \mathbb{R}^{M\times C}$ and durations $\Tilde{\mathbf{a}}^t \in \mathbb{R}^{M\times 1}$ through two dedicated linear layers. To guarantee the non-negativity of the predicted durations, an exponential activation function is employed.

\myparagraph{Inference.} During inference, future action embeddings $\mathbf{z}_S$ are drawn from a standard Gaussian distribution $\mathcal{N}(\boldsymbol{0}, \boldsymbol{I})$. Alternatively, we can initialize $\mathbf{z}_S$ with zero vectors (mean of the standard Gaussian distribution) for deterministic results. The decoder is made step-aware by embedding the current diffusion step $s=S$ with sinusoidal positional encodings, futher integrated via a multi-layer perceptron.
The decoder inputs $\mathbf{z}_S$, $Q$, and step data, aiming to initially rectify the noisy futures and predict $\Tilde{\mathbf{z}}_0$, taking cues from the encoded past observations $E$ via cross-attention. For $E$ that does not match the decoder's dimension $D'$, a linear transformation adjusts its dimensionality. To enable the iterative reverse diffusion process, we apply the forward chain to get $\Tilde{\mathbf{z}}_{S-1}$ and input it with the updated step $s=S-1$ to the decoder. By iteratively removing noise, a final sample $\Tilde{\mathbf{z}}_0$ is generated via a trajectory $\mathbfz_{S}, \Tilde{\mathbfz}_{S-1}, ..., \Tilde{\mathbf{z}}_0$. To accelerate the reverse process, we adopt DDIM~\cite{song2021ddim} to skip steps in the trajectory. The resulting action embeddings $\Tilde{\mathbf{z}}_0$ are decoded into future action labels and durations by the predictor. 

Given the preset number of action queries $M$ may surpass the actual number of future actions $N$, an \texttt{EOS} class is introduced to signify sequence termination. 
In the inference stage, following~\cite{gongFutureTransformerLongterm2022}, all predictions after the first prediction of \texttt{EOS} are discarded. Moreover, we apply Gaussian normalization to the predicted durations to make the sum of all durations equates to one, consistent with the previous work~\cite{farhaWhenWillYou2018,gongFutureTransformerLongterm2022}.

\subsection{Training}
\label{sec:training}
\begin{figure}[t]
    \centering
    \includegraphics[width=\linewidth]{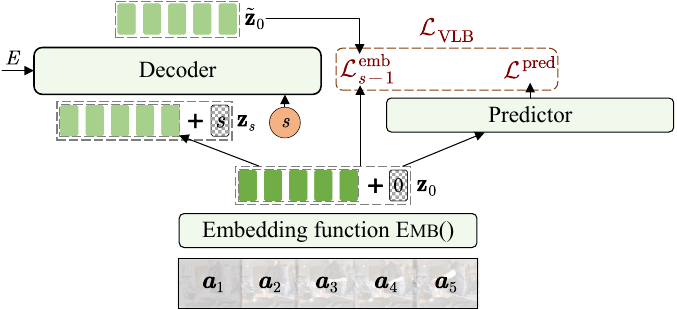}
    \caption{Training pipeline. By employing an embedding function and an action predictor, we integrate the discrete future actions $\boldsymbol{a}$ into the standard diffusion models. The training optimizes a variational lower bound objective, which combines an embedding reconstruction loss $\mathcal{L}^{\text{emb}}_{s-1}$ and an action prediction loss $\mathcal{L}^{\text{pred}}$.}
    \label{fig:training}
\end{figure}

To integrate discrete actions $\boldsymbol{a}$ into the continuous diffusion processes, we extend standard diffusion models with a future action embedding function and an action predictor to facilitate the conversion between continuous embeddings $\mathbf{z}_0$ and discrete actions $\boldsymbol{a}$, as illustrated in Fig.~\ref{fig:training}. The following details \model's learning process.  

\myparagraph{Forward Process with Future Actions as Input.}
\model takes future actions $\boldsymbol{a}$, which comprise action classes $\boldsymbol{a}^c \in \mathbb{R}^{M \times C}$ and durations $\boldsymbol{a}^t \in \mathbb{R}^{M \times 1}$, where $C$ is the total number of action classes. It first linearly transforms the one-hot encoded classes and real-value durations into embeddings, $\textsc{Emb}_c (\boldsymbol{a}^c)$ and $\textsc{Emb}_t (\boldsymbol{a}^t)$, respectively. An additional linear layer then merges these embeddings to form a joint feature space $\textsc{Emb}((\boldsymbol{a}^c, \boldsymbol{a}^t)) = f_a([\textsc{Emb}_c (\boldsymbol{a}^c), \textsc{Emb}_t (\boldsymbol{a}^t)])$. 
When only action classes are predicted, the model bypasses $f_a$ and directly concatenates the class embeddings.
This step facilitates the inclusion of discrete future actions into the diffusion model's forward process by extending the original forward chain to a new Markov transition ${q_{\phi}}(\mathbf{z}_{0}|\boldsymbol{a})=\mathcal{N}(\textsc{Emb}(\boldsymbol{a}), \beta_0 \boldsymbol{I})$, as shown in Fig.~\ref{fig:concept}.

\myparagraph{Reverse Process with Conditional Denoising.}
The reverse process aims to recover the initial state $\mathbf{z}_0$ from the noised version $\mathbf{z}_S$ by applying the denoising probability $p_{\theta}(\mathbf{z}_{0:S}):=p(\mathbf{z}_S)\prod_{s=1}^Sp_{\theta}(\mathbf{z}_{s-1}|\mathbf{z}_s)$. Incorporating encoded past observations $E$ as conditional inputs transforms the learning process into $p_{\theta}(\mathbf{z}_{s-1}|\mathbf{z}_s, E)$, modeled with the proposed \model $f_{\theta}(\mathbf{z}_s, s,E)$.
For training, in line with~\cite{li2022diffusionlm,gong2023diffuseq}, we streamline the variational lower bound ($\mathcal{L}_\text{VLB}$) to consist of an embedding reconstruction loss and an action prediction loss, $\mathcal{L}_{\text{VLB}} = \sum_{s=1}^S \mathcal{L}^{\text{emb}}_{s-1} + \mathcal{L}^{\text{pred}}$. The embedding loss $\mathcal{L}^{\text{emb}}_{s-1}$ is calculated by:
\begin{equation}
    \mathcal{L}^{\text{emb}}_{s-1} = 
\begin{cases} 
||\mathbf{z}_0 - f_{\theta}(\mathbf{z}_s, s,E)||^2 & \text{if } 2 \!\leq\! s \!\leq\! S \\
||\textsc{Emb}(\boldsymbol{a}) - f_{\theta}(\mathbf{z}_1, 1,E)||^2 & \text{if } s = 1
\end{cases}
\end{equation}
and the prediction loss $\mathcal{L}^{\text{pred}}$ is defined as: 
$\mathcal{L}^{\text{pred}} = - \log p_{\theta}(\boldsymbol{a}^c|\mathbf{z}_0,E) - \log p_{\theta}(\boldsymbol{a}^t|\mathbf{z}_0,E)$, under the assumption that action classes and durations are conditionally independent for mathematical convenience, following common practices~\cite{mehrasaVariationalAutoEncoderModel2019}.
The action class log-likelihood is computed using cross-entropy and the duration log-likelihood is computed by the mean squared error (MSE) loss. More details are provided in the supplementary material.

\myparagraph{Training Objective.}
In addition to $\mathcal{L}_\text{VLB}$ for the decoder outputs, we append a classification head to the encoder and apply a cross-entropy loss $\mathcal{L}_{\text{seg}}$ and a temporal smoothness loss $\mathcal{L}_{\text{smooth}}$ as auxiliary supervision. The temporal smoothness loss~\cite{farha2019ms-tcn} is computed as the mean squared error of the log-likelihoods between adjacent video frames to promote the local similarity along the temporal dimension. The final training objective is thus defined as
\begin{equation}
    \mathcal{L} = \mathcal{L}_{\text{seg}} + \mathcal{L}_{\text{smooth}} + \mathcal{L}_\text{VLB}.
\end{equation}

\begin{table*}[ht!]
    \setlength\tabcolsep{3pt}
    \resizebox{\linewidth}{!}{
    \begin{tabular}{@{}cclcccccccccccccccc@{}}
    \toprule
    \multirow{2}{*}{\rotatebox{90}{Type}}& \multirow{2}{*}{\shortstack{Back- \\ bone}} & \multirow{2}{*}{Methods} & \multicolumn{4}{c}{Breakfast $\beta$ ($\alpha$ = 0.2)} & \multicolumn{4}{c}{Breakfast $\beta$ ($\alpha$ = 0.3)} & \multicolumn{4}{c}{50Salads $\beta$ ($\alpha$ = 0.2)} & \multicolumn{4}{c}{50Salads $\beta$ ($\alpha$ = 0.3)} \\
    \cmidrule(lr){4-7} \cmidrule(lr){8-11} \cmidrule(lr){12-15} \cmidrule(lr){16-19}
    & & & 0.1 & 0.2 & 0.3 & 0.5 & 0.1 & 0.2 & 0.3 & 0.5 & 0.1 & 0.2 & 0.3 & 0.5 & 0.1 & 0.2 & 0.3 & 0.5\\
    \midrule
    \multirow{4}{*}{\rotatebox{90}{Pred. label}} & \multirow{4}{*}{Fisher} & RNN~\cite{farhaWhenWillYou2018} & 18.11 & 17.20 & 15.94 & 15.81 & 21.64 & 20.02 & 19.73 & 19.21 & 30.06 & 25.43 & 18.74 & 13.49 & 30.77 & 17.19 & 14.79 & 9.77 \\
    & & CNN~\cite{farhaWhenWillYou2018}  & 17.90 & 16.35 & 15.37 & 14.54 & 22.44 & 20.12 & 19.69 & 18.76 & 21.24 & 19.03 & 15.98 & 9.87 & 29.14 & 20.14 & 17.46 & 10.86 \\
    & & UAAA~\cite{abufarhaUncertaintyAwareAnticipationActivities2019} & 16.71 & 15.40 & 14.47 & 14.20 & 20.73 & 18.27 & 18.42 & 16.86 & 24.86 & 22.37 & 19.88 & 12.82 & 29.10 & 20.50 & 15.28 & 12.31 \\
    & & Timecond.~\cite{keTimeConditionedActionAnticipation2019}  & 18.41 & 17.21 & 16.42 & 15.84 & 22.75 & 20.44 & 19.64 & 19.75 & 32.51 & 27.61 & 21.26 & 15.99 & 35.12 & 27.05 & 22.05 & 15.59 \\

     \midrule
     
     \multirow{7}{*}{\rotatebox{90}{Features}} & \multirow{2}{*}{Fisher}  & CNN~\cite{farhaWhenWillYou2018} & 12.78 & 11.62 & 11.21 & 10.27 & 17.72 & 16.87 & 15.48 & 14.09 & -- & -- & -- & -- & -- & -- & -- & -- \\
    & & TempAgg~\cite{senerTemporalAggregateRepresentations2020} & 15.60 & 13.10 & 12.10 & 11.10 & 19.50 & 17.00 & 15.60 & 15.10 & 25.50 & 19.90 & 18.20 & 15.10 & 30.60 & 22.50 & 19.10 & 11.20 \\[.4mm]
     
    \arrayrulecolor{lightgray}\hhline{*{1}{~}*{18}{-}}\arrayrulecolor{black} \rule{0pt}{2.4ex}
    & \multirow{5}{*}{I3D} & 
    TempAgg~\cite{senerTemporalAggregateRepresentations2020} & 24.20 & 21.10 & 20.00 & 18.10 & 30.40 & 26.30 & 23.80 & 21.20 & -- & -- & -- & -- & -- & -- & -- & -- \\
    & & Cycle Cons\cite{abu2020long}  & 25.88 & 23.42 & 22.42 & 21.54 & 29.66 & 27.37 & 25.58 & 25.20 & 34.76 & 28.41 & 21.82 & 15.25 & 34.39 & 23.70 & 18.95 & 15.89 \\
    & & A-ACT~\cite{gupta2022act} & \underline{26.70} & 24.30 & \underline{23.20} & 21.70 & 30.80 & 28.30 & 26.10 & 25.80 & 35.40 & \underline{29.60} & 22.50 & 16.10 & \textbf{35.70} & \underline{25.30} & 20.10 & \underline{16.30} \\
    & & FUTR~\cite{gongFutureTransformerLongterm2022} & \textbf{27.70} & \underline{24.55} & 22.83 & \underline{22.04} & \textbf{32.27} & \underline{29.88} & \underline{27.49} & \underline{25.87} & \textbf{39.55} & 27.54 & \underline{23.31} & \underline{17.77} & \underline{35.15} & 24.86 & \underline{24.22} & 15.26 \\ 

    \arrayrulecolor{lightgray}\hhline{*{2}{~}*{17}{-}}\arrayrulecolor{black} \rule{0pt}{2.4ex}

    & & \model & 25.33 & \textbf{24.59} & \textbf{24.39} & \textbf{22.74} & \underline{32.13} & \textbf{31.83} & \textbf{31.18} & \textbf{30.77} & \underline{36.13} & \textbf{34.00} & \textbf{30.46} & \textbf{25.29} & 34.09 & \textbf{30.14} & \textbf{26.34} & \textbf{20.23} \\
    
     \bottomrule
    \end{tabular}}
    \caption{Comparison with state-of-the-art methods on Breakfast~\cite{kuehneLanguageActionsRecovering2014} and 50Salads~\cite{stein2013combining} using MoC (\%). Bold and underlined numbers indicate the highest and second highest accuracy, respectively. \model achieves the state-of-the-art performance in almost all settings.}
    \label{tab:sota_breakfast_50salads}
\end{table*}

\section{Experimental Setup}

\noindent\textbf{Datasets.}
The Breakfast~\cite{kuehneLanguageActionsRecovering2014} dataset comprises 1,712 videos of 52 different individuals making breakfast in 18 different kitchens, totalling 77 hours. Every video is categorized into one of the 10 activities related to breakfast preparation.
The videos are annotated by 48 fine-grained actions.  
The 50Salads~\cite{stein2013combining} dataset comprises 50 top-view videos of 25 people preparing a salad. The dataset contains over 4 hours of RGB-D video data, annotated with 17 fine-grained action labels and 3 high-level activities.
EpicKitchens~\cite{damen2018scaling} and EGTEA~Gaze+~\cite{li2018eye} are egocentric datasets.
EpicKitchens contains 39,596 segments labeled with 125 verbs, 352 nouns, and 2,513 combinations (actions), totalling 55 hours. EGTEA Gaze+ contains 28 hours of videos including 10.3K action annotations, 19 verbs, 51 nouns, and 106 unique actions.

\myparagraph{Metrics.} For both Breakfast and 50Salads datasets, we calculate the mean accuracy over classes, averaged across all future timestamps within a specified anticipation duration. We observe the first portion ($\alpha$) of a video, adhering to benchmarks from~\cite{farhaWhenWillYou2018} that set $\alpha$ values at 0.2 or 0.3. Subsequent predictions cover segments $\beta$ of the entire video, where $\beta$ can be one of \{0.1, 0.2, 0.3, 0.5\}. For our evaluation metrics, we average results across 4 splits for the Breakfast dataset and 5 splits for the 50Salads dataset.
For the EpicKitchens and EGTEA Gaze+ datasets, we employ a multi-label classification metric (mAP) targeting specific action classes. A portion $\alpha$ of each untrimmed video serves as input to forecast all subsequent action classes, representing the remaining ($1 - \alpha$) duration of the video. As in~\cite{nagarajan2020ego}, we use $\alpha=\{0.25, 0.50, 0.75\}$ for evaluation.
In alignment with~\cite{nagarajan2020ego}, we also report the mAP metrics in both low-shot (rare) and many-shot (freq) scenarios.

\myparagraph{Architecture Details.}
The encoder has four layers with a hidden dimension $D=256$. The decoder has 8 layers for 50Salads and 4 layers for the other datasets. The decoder dimensions $D'$ are set at 1024 for Breakfast and EpicKitchens, 512 for EGTEA Gaze+, and 256 for 50Salads. We set the number of action queries $M$ to 8 for Breakfast and 16 for 50Salads, since 50Salads includes more actions than Breakfast in a video. For EpicKitchens and EGTEA Gaze+, we set $M$ to 1, as the task is treated as a multi-label task defined in~\cite{nagarajan2020ego}. 

\myparagraph{Training Details.}
To allow a fair comparison to other state-of-the-art long-term action anticipation methods, we use the pre-extracted I3D features~\cite{carreiraQuoVadisAction2017} as input visual features $F$ for all datasets, provided by~\cite{farha2019ms-tcn} and~\cite{nagarajan2020ego}. We sample the I3D features with a stride of 3 for Breakfast and 50Salads and 1 for EpicKitchens and EGTEA Gaze+. In training, we set the observation rate $\alpha \in$ \{0.2, 0.3, 0.4, 0.5\} for Breakfast and 50Salads and additionally use \{0.6, 0.7, 0.8\} for EpicKitchens and EGTEA. We use AdamW optimizer and train our model for 50, 30, 50, 100 epochs for Breakfast, 50Salads, EpicKitchens, and EGTEA Gaze+, respectively. We employ a cosine annealing warm-up scheduler with warm-up stages of 10 epochs. The total number of steps is set as $S=1000$. To sample diffusion steps $s$, we adopt importance sampling~\cite{nichol2021improved,gong2023diffuseq} in our experiments. 
For all trained models, we apply gradient clipping at 1. 
More training details can be found in the supplementary material.

\section{Results}
To compare \model with the state-of-the-art methods in Section~\ref{sec:sota} and conduct ablation studies in Section~\ref{sec:ablation}, we initialize the action embeddings $\mathbf{z}_S$ as zeros for deterministic predictions. In Section~\ref{sec:diversity} and Section~\ref{sec:intermediate}, we sample $\mathbf{z}_S$ from the standard Gaussian distribution, and present an in-depth analysis of our approach.

\subsection{Comparison with the State-of-the-art}
\label{sec:sota}
In Table~\ref{tab:sota_breakfast_50salads}, we compare our methods with the state-of-the-art methods on Breakfast and 50Salads.
While we list the methods that take the action labels predicted by an action segmentation method (\eg, \cite{richard2017weakly}) at the top of the table for completeness, we mainly compare our method with methods that take visual features as input. Note that we do not list the results of \textsc{Anticipatr}~\cite{nawhal2022rethinking} in Table~\ref{tab:sota_breakfast_50salads}, as they used a different evaluation protocol~\cite{zhong2023survey}.
\model achieves the state-of-the-art performance in almost all experimental settings on Breakfast and 50Salads. Notably, our method outperforms other methods with a large margin for long time horizons for anticipation, \eg, for $\alpha=0.3$ and $\beta=0.5$, accuracy improvement on Breakfast and 50Salads is 18.9\% (25.87 $\rightarrow$ 30.77) and 24.1\% (16.30 $\rightarrow$ 20.23), respectively.

\begin{table}[t]
    \centering
    \setlength\tabcolsep{3pt}
    \resizebox{0.9\linewidth}{!}{
    \begin{tabular}{@{}lcccccc@{}}
        \toprule
        \multirow{2}{*}{Method} & \multicolumn{3}{c}{EpicKitchens} & \multicolumn{3}{c}{EGTEA Gaze+} \\
        \cmidrule(lr){2-4} \cmidrule(lr){5-7}  
        & All & Freq & Rare & All & Freq & Rare \\ \midrule
        I3D~\cite{carreiraQuoVadisAction2017} & 32.7 & 53.3 & 23.0 & 72.1 & 79.3 & 53.3  \\
        ActionVLAD~\cite{girdhar2017actionvlad} & 29.8 & 53.5 & 18.6 & 73.3 & 79.0 & 58.6 \\ 
        Timeception~\cite{hussein2019timeception} & 35.6 & 55.9 & 26.1 & 74.1 & 79.7 & \underline{59.7} \\ 
        VideoGraph~\cite{hussein2019videograph} & 22.5 & 49.4 & 14.0 & 67.7 & 77.1 & 47.2 \\ 
        EGO-TOPO~\cite{nagarajan2020ego} & 38.0 & \underline{56.9} & \underline{29.2} & 73.5 & 80.7 & 54.7 \\ 
        \textsc{Anticipatr}~\cite{nawhal2022rethinking} & \textbf{39.1} & \textbf{58.1} & 29.1 & \underline{76.8} & \underline{83.3} & 55.1 \\
        
        \arrayrulecolor{lightgray}\hhline{{~}*{7}{-}}\arrayrulecolor{black} \rule{0pt}{2.4ex}\model (Ours) & \underline{38.7} & 55.0 & \textbf{31.0} & \textbf{77.3} & \textbf{83.5} & \textbf{61.4}\\
        \bottomrule
    \end{tabular}}
    \caption{Comparison with state-of-the-art methods on EpicKitchens~\cite{damen2018scaling} and EGTEA Gaze+~\cite{li2018eye} in mAP. Bold and underlined numbers indicate the highest and second highest accuracy, respectively. \model achieves competitive results on EpicKitchens and sets a new state-of-the-art on EGTEA Gaze+.}
    \label{tab:sota_ek55_egtea}
\end{table}

Next we evaluate our method on EpicKitchens and EGTEA Gaze+, as presented in Table~\ref{tab:sota_ek55_egtea}. The results indicate that our approach achieves competitive results when compared to the state-of-the-art method~\cite{nawhal2022rethinking} on EpicKitchens. While our method ranks second for the \textit{All} set, it achieves the best results for the more challenging \textit{Rare} set. Notably, our approach does not require separate training for the encoder and decoder, a step that is necessary in~\cite{nawhal2022rethinking}. For EGTEA Gaze+, our method establishes a new benchmark, setting the state-of-the-art across all sets.

\begin{figure*}[ht!]
    \begin{minipage}{0.3\linewidth}
        \centering
        Observation
    \end{minipage}
    \begin{minipage}{0.6\linewidth}
        \centering
        Prediction
    \end{minipage}
    
    \centering
    \includegraphics[width=0.95\linewidth]{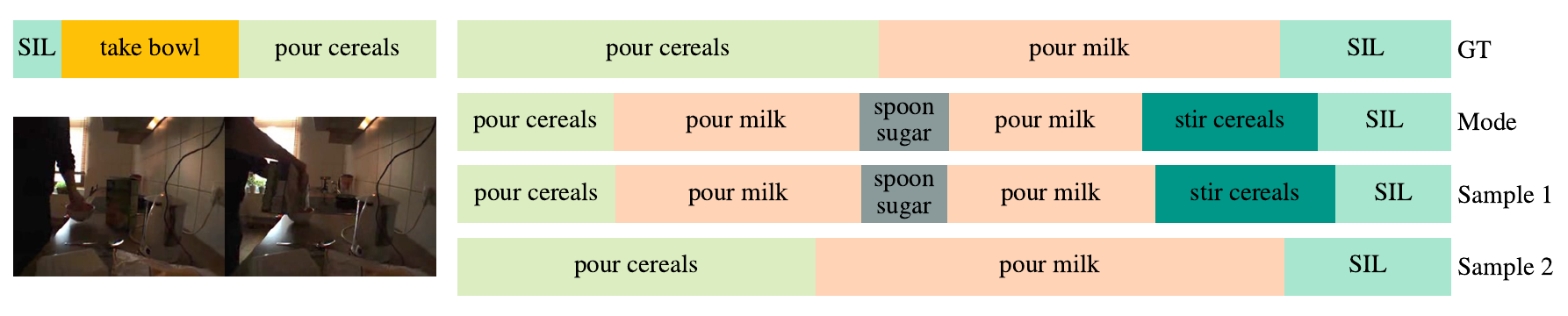}\vspace{-0.2cm}
    \caption{Qualitative results on Breakfast. Observations are shown on the left, while the ground-truth labels and predicted results, \ie, one deterministic prediction and two randomly sampled results, are displayed on the right. We set $\alpha$ as 0.3 and predict all subsequent actions in this experiment. Action labels and durations are decoded as frame-wise action classes. More qualitative results are provided in the supplementary material.}
    \label{fig:visualization}
\end{figure*}

\subsection{Ablation Study}
\label{sec:ablation}
Extensive ablation studies are performed to validate the
design choices in our method. The experiments on Breakfast are conducted across all splits, and the results are averaged across these splits.
Default settings are marked in \colorbox{Gray}{gray}.

\begin{table}[t]
    \centering
    \setlength\tabcolsep{3pt}
    \resizebox{0.85\linewidth}{!}{
    \begin{tabular}{lccccc}
        \toprule
        \multirow{2}{*}{Encoder} & \multicolumn{4}{c}{$\beta$ ($\alpha$ = 0.3)} \\
        \cmidrule(lr){2-5} 
        & 0.1 & 0.2 & 0.3 & 0.5 \\
        \midrule
         \rowcolor{Gray} DETR Encoder~\cite{carion2020end} & \textbf{32.13} & \textbf{31.83} & \textbf{31.18} & \textbf{30.77} \\
         DETR Encoder~\cite{carion2020end} (local) & 31.01 & 30.58 & 30.38 & 29.16 \\
         RoFormer~\cite{su2021roformer} & 22.26 & 21.94 & 22.42 & 21.20 \\
         \bottomrule
    \end{tabular}}
    \caption{Impact of the encoder architecture on Breakfast in MoC.}
    \label{tab:ablation_encoder}
\end{table}

\myparagraph{Encoder Architecture.}
We first evaluate the impact of different encoder architectures in Table~\ref{tab:ablation_encoder}.  
In our default setting, we follow~\cite{carion2020end,gongFutureTransformerLongterm2022} to use a transformer encoder with global attention across all past frames. As locality of features has been found particularly beneficial for action segmentation~\cite{yi2021asformer}, we similarly apply a hierarchical attention mask to the attention matrix, forcing the shallow layers to concentrate on neighboring frames, while deeper layers are allowed to attend to global frames. Details can be found in the supplementary material. Additionally, we employ the rotary position embedding technique~\cite{su2021roformer}, which allows to endow the transformer with relative positional embeddings without learnable parameters and has been widely adopted in the NLP community.
The default setting consistently outperforms the other architectures across all prediction horizons ($\beta$), highlighting its efficacy in this context.

\begin{table}[t]
    \centering
    \setlength\tabcolsep{3pt}
    \resizebox{0.88\linewidth}{!}{
    \begin{tabular}{@{}cccccccc@{}}
        \toprule
        \multicolumn{3}{c}{Loss} & \multicolumn{4}{c}{Breakfast $\beta$ ($\alpha$ = 0.3)} & EGTEA \\
        \cmidrule(lr){1-3} \cmidrule(lr){4-7} \cmidrule(lr){8-8} 
        $\mathcal{L}_\text{seg}$ & $\mathcal{L}_\text{smooth}$ & $\mathcal{L}_\text{VLB}$ & 0.1 & 0.2 & 0.3 & 0.5 & All \\
        \midrule
         \xmark & \xmark & \checkmark & 19.19 & 18.31 & 17.70 & 16.14 & 74.11 \\
         \checkmark & \xmark & \checkmark & 31.66 & 31.44 & \textbf{32.17} & \textbf{30.91} & 77.07 \\
         \rowcolor{Gray} \checkmark & \checkmark & \checkmark & \textbf{32.13} & \textbf{31.83} & 31.18 & 30.77 & \textbf{77.33} \\
         \bottomrule
    \end{tabular}}
    \caption{Impact of loss terms. The action segmentation loss $\mathcal{L}_\text{seg}$ substantially improves the anticipation performance, indicating the importance of recognizing past frames in effectively anticipating future actions. The temporal smoothness loss $\mathcal{L}_\text{smooth}$ is optional and slightly improves the results, but not for all settings. 
    }
    \label{tab:ablation_loss}
\end{table}
\begin{table}[t]
    \centering
    \begin{subtable}[t]{0.474\linewidth}
    \setlength\tabcolsep{2pt}
    \resizebox{\linewidth}{!}{
    \begin{tabular}{cccccc}
        \toprule
        \multirow{2}{*}{$M$} & \multicolumn{4}{c}{Breakfast $\beta$ ($\alpha$ = 0.3)} \\
        \cmidrule(lr){2-5} 
        & 0.1 & 0.2 & 0.3 & 0.5 \\
        \midrule
         6 & 31.42 & 31.20 & \textbf{31.38} & 30.57 \\
         \rowcolor{Gray} 8 & \textbf{32.13} & \textbf{31.83} & 31.18 & \textbf{30.77} \\
         10 & 30.50 & 30.01 & 30.43 & 28.57 \\
         12 & 29.18 & 28.28 & 29.39 & 26.88 \\
         \bottomrule
    \end{tabular}}
    \subcaption{\# Action queries.}
    \label{tab:ablation_queries}
    \end{subtable}
    \begin{subtable}[t]{0.516\linewidth}
    \setlength\tabcolsep{2pt}
    \resizebox{\linewidth}{!}{
        \begin{tabular}{cccccc}
        \toprule
        \multirow{2}{*}{Steps} & \multicolumn{4}{c}{Breakfast $\beta$ ($\alpha$ = 0.3)} \\
        \cmidrule(lr){2-5} 
        & 0.1 & 0.2 & 0.3 & 0.5 \\
        \midrule
         25 & \textbf{32.32} & 31.81 & 33.05 & 30.14 \\
         50 & 31.90 & 31.46 & \textbf{33.14} & 30.09 \\
         \rowcolor{Gray} 100 & 32.13 & \textbf{31.83} & 31.18 & \textbf{30.77} \\
         200 & 31.41 & 31.02 & 31.38 & 30.11 \\
         \bottomrule
    \end{tabular}}
    \subcaption{\# Inference steps.}
    \label{tab:ablation_steps}
    \end{subtable}
    
    \caption{Ablation study on number of action queries $M$ and inference steps on Breakfast. (a) Using $M=8$ queries is sufficient. (b) 100 inference steps are sufficient.}
    \label{tab:ablation_queries_steps}
\end{table}

\myparagraph{Loss Terms.}
In Table~\ref{tab:ablation_loss}, we assess the impact of the action segmentation loss and the temporal smoothness loss on Breakfast and EGTEA Gaze+. As also observed in~\cite{gongFutureTransformerLongterm2022}, the action segmentation loss substantially enhances the anticipation performance. This underscores the importance of recognizing past frames to effectively anticipate future actions. While introducing the temporal smoothness loss into the framework improves the performance on EGTEA Gaze+, its effect on Breakfast is not clearly evident. 

\myparagraph{Number of Action Queries.}
In Table~\ref{tab:ablation_queries}, we analyze the impact of the number of action queries $M$. We vary $M$, starting from 6 and increasing in increments of 2 up to 12. For the Breakfast dataset, 6-8 queries are enough. Using more than 8 queries decreases the accuracy.

\myparagraph{Diffusion Inference Steps.}
We adopt DDIM~\cite{song2021ddim} to skip inference steps in our work. We vary the number of total inference steps and report the results in Table~\ref{tab:ablation_steps}. Using 100 iterations performs for most settings best, but even 25 iterations delivers reasonable results.    

\subsection{Uncertainty-Aware Anticipation}
\label{sec:diversity}

\begin{table}[t]
    \centering
    \setlength\tabcolsep{2pt}
    \resizebox{\linewidth}{!}{
    \begin{tabular}{@{}cccccccccc@{}}
        \toprule
        \multirow{2}{*}{Method} & \multirow{2}{*}{$m$} & \multicolumn{4}{c}{$\beta$ ($\alpha$ = 0.2)} & \multicolumn{4}{c}{$\beta$ ($\alpha$ = 0.3)} \\
        \cmidrule(lr){3-6} \cmidrule(lr){7-10} 
        & & 0.1 & 0.2 & 0.3 & 0.5 & 0.1 & 0.2 & 0.3 & 0.5 \\
        \midrule
         \cite{abufarhaUncertaintyAwareAnticipationActivities2019} & 25 & 15.69 & 14.00 & 13.30 & 12.95 & 19.14 & 17.18 & 17.38 & 14.98 \\
         Ours & 25 & \textbf{24.74} & \textbf{22.89} & \textbf{22.09} & \textbf{22.34} & \textbf{30.92} & \textbf{30.17} & \textbf{28.85} & \textbf{27.46} \\
         \arrayrulecolor{lightgray}\hline\arrayrulecolor{black} \rule{0pt}{2.3ex}\multirow{6}{*}{Ours} & 5 & 25.10 & 22.80 & 22.00 & 21.07 & 30.83 & 29.64 & 28.32 & 25.88 \\
         & 10 & \textbf{24.94} & \textbf{23.34} & \textbf{22.74} & 21.78 & 30.86 & 29.31 & 28.66 & 26.66 \\
         & 25 & 24.74 & 22.89 & 22.09 & 22.34 & 30.92 & 30.17 & \textbf{28.85} & \textbf{27.46} \\
         & 50 & 24.81 & 23.24 & 22.26 & \textbf{22.67} & 30.92 & \textbf{30.37} & 28.80 & 27.22 \\
         & 100 & 24.77 & 23.29 & 22.42 & 22.18 & \textbf{30.98} & 30.05 & 28.67 & 27.40 \\
         \bottomrule
    \end{tabular}}
    \caption{Diverse anticipation performance on Breakfast averaged over $m$ random samples for each observation. \model significantly exceeds the probabilistic approach~\cite{abufarhaUncertaintyAwareAnticipationActivities2019}.}
    \label{tab:evaluation_diversity_full}
\end{table}
\begin{table}[t]
    \centering
    \setlength\tabcolsep{2pt}
    \resizebox{\linewidth}{!}{
    \begin{tabular}{@{}cccccccccc@{}}
        \toprule
        \multirow{2}{*}{Method} & \multirow{2}{*}{$m$} & \multicolumn{4}{c}{$\beta$ ($\alpha$ = 0.2)} & \multicolumn{4}{c}{$\beta$ ($\alpha$ = 0.3)} \\
        \cmidrule(lr){3-6} \cmidrule(lr){7-10} 
        & & 0.1 & 0.2 & 0.3 & 0.5 & 0.1 & 0.2 & 0.3 & 0.5 \\
        \midrule
        
        \cite{abufarhaUncertaintyAwareAnticipationActivities2019} & 25 & 28.89 & 28.43 & 27.61 & 28.04 & 32.38 & 31.60 & 32.83 & 30.79 \\
        Ours & 25 & \textbf{31.27} & \textbf{29.87} & \textbf{29.40} & \textbf{30.11} & \textbf{37.38} & \textbf{37.01} & \textbf{36.29} & \textbf{34.80} \\

        \arrayrulecolor{lightgray}\hline\arrayrulecolor{black} \rule{0pt}{2.3ex}
        
         \multirow{6}{*}{Ours} & -- & 25.33 & 24.59 & 24.39 & 22.74 & 32.13 & 31.83 & 31.18 & 30.77 \\
         \arrayrulecolor{lightgray}\hhline{*{1}{~}*{9}{-}}\arrayrulecolor{black} \rule{0pt}{2.3ex}
         & 5 & 27.62 & 25.32 & 24.79 & 24.37 & 35.12 & 32.76 & 31.11 & 29.27 \\
         & 10 & 28.77 & 27.49 & 27.36 & 27.16 & 35.63 & 34.16 & 32.96 & 31.59 \\
         & 25 & 31.27 & 29.87 & 29.40 & 30.11 & 37.38 & 37.01 & 36.29 & 34.80 \\
         & 50 & 31.08 & 30.54 & 30.34 & 31.52 & 37.35 & 38.13 & 37.51 & 36.22 \\
         & 100 & \textbf{32.43} & \textbf{32.00} & \textbf{31.87} & \textbf{33.27} & \textbf{38.26} & \textbf{39.10} & \textbf{39.79} & \textbf{38.54} \\
         \bottomrule
    \end{tabular}}
    \caption{Diverse anticipation performance on Breakfast using top-1 accuracy, \ie, accuracy of best match to ground truth of $m$ random samples for each observation. \model surpasses the probabilistic approach~\cite{abufarhaUncertaintyAwareAnticipationActivities2019}. -- denotes the result of our approach in the deterministic setting.}
    
    \label{tab:diversity_best_match_full}
\end{table}

\begin{figure*}[t]
    \begin{minipage}{0.3\linewidth}
        \centering
        Observation
    \end{minipage}
    \begin{minipage}{0.6\linewidth}
        \centering
        Prediction
    \end{minipage}
    
    \centering
    \includegraphics[width=0.95\linewidth]{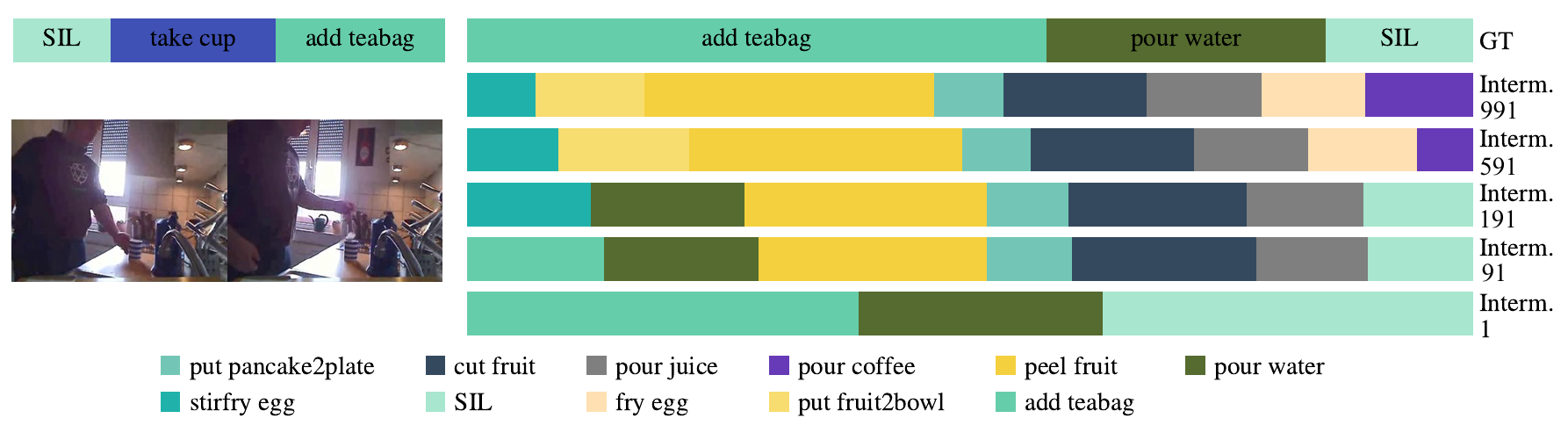}\vspace{-0.3cm}
    \caption{Anticipation results of the intermediate inference steps on Breakfast. Observations are shown on the left, while the ground-truth labels and predicted results are displayed on the right. $\alpha$ is set as 0.3 and all subsequent actions are predicted in this experiment. As the inference step approaches 1, the predictions are gradually refined, and the ground truth actions emerge.}
    \label{fig:intermediate_diffusion}
\end{figure*}

To evaluate the quality of our method for non-deterministic anticipation, we use the protocols that have been proposed in~\cite{abufarhaUncertaintyAwareAnticipationActivities2019}. The protocols require the generation of $m$ future predictions for the same video observation, calculating either the average accuracy of all generated predictions or the top-1 accuracy. We first report the results for the averaging protocol.  
For our method, we generate $m$ future predictions by randomly sampling $m$ future action embeddings $\mathbf{z}_S$ from the Gaussian distribution $\mathcal{N}(\boldsymbol{0}, \boldsymbol{I})$. We then compute the average performance of these samples across all splits of Breakfast. Since the previous probabilistic generative methods~\cite{mehrasaVariationalAutoEncoderModel2019,zhao2020diverse} rely on ground truth action labels, we only compare our approach with the probabilistic approach~\cite{abufarhaUncertaintyAwareAnticipationActivities2019}.
The results for varying $m$ are presented in Table~\ref{tab:evaluation_diversity_full}, demonstrating that the averaged performance of \model maintains relatively stable with respect to changes in $m$. Moreover, our method significantly outperforms the approach~\cite{abufarhaUncertaintyAwareAnticipationActivities2019} if we use the same number of samples $m=25$.

It is important to recognize that the provided ground truth future in the dataset represents merely one among several feasible futures. This is exemplified in Fig.~\ref{fig:visualization}, where the left side depicts the observations, and the right side shows the ground truth labels alongside three predictions from \model. We label the prediction for zero $\mathbf{z}_S$ as \textit{Mode} and two randomly sampled $\mathbf{z}_S$ as \textit{Sample 1} and \textit{Sample 2}. Notably, while Sample 2 matches with the provided ground truth, the other predictions also appear logical.
Consequently, we adopt the top-1 protocol proposed in~\cite{abufarhaUncertaintyAwareAnticipationActivities2019}, which focuses on the performance of the prediction that best aligns with the ground truth, acknowledging that the ground truth represents just one of several possible futures. Specifically, for $m$ generated predictions for the same observation, we select the prediction with the highest number of correctly predicted future frames. The performance for different $m$ values is presented in Table~\ref{tab:diversity_best_match_full}. For comparison, we also include results where $\mathbf{z}_S$ is a zero vector (denoted with -- for $m$). The performance of our non-deterministic anticipations consistently surpasses that of our deterministic one and the probabilistic approach~\cite{abufarhaUncertaintyAwareAnticipationActivities2019} with few samples, underscoring the superior quality of \model. Furthermore, as expected, we observe a consistent increase in performance as the number of samples grows.

\begin{figure}[t]
    \centering
    \includegraphics[width=0.85\linewidth]{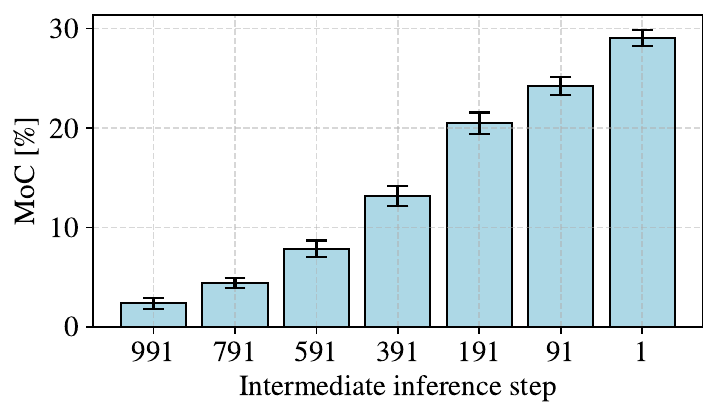}\vspace{-0.2cm}
    \caption{Intermediate anticipation performance on Breakfast. We run our model 25 times for each split and show the overall mean and standard deviation for the intermediate inference steps. $\alpha$ and $\beta$ are set as 0.3 and 0.5, respectively.}
    \label{fig:intermediate_plot}
\end{figure}

\subsection{Intermediate Diffusion Anticipation}
\label{sec:intermediate}

In this section, we delve into the denoising capability of the proposed method and assess the anticipation performance of the intermediate inference steps. As the total number of diffusion steps is set as $S=1000$, with 100 steps used in inference, the inference trajectory becomes $991, 981, ..., 1$. Fig.~\ref{fig:intermediate_plot} provides a quantitative perspective across these varying intermediate inference steps. For this evaluation, we rerun our model 25 times and compute the performance of each run for the randomly sampled $\mathbf{z}_S$. We then calculate the mean and standard deviation of these runs for each particular inference step.
A discernible trend emerges from the figure: as the inference step approaches 1, there is a marked rise in the performance. This upward trajectory signifies the enhanced accuracy during the reverse diffusion process. 

Fig.~\ref{fig:intermediate_diffusion} presents the anticipation results derived at various intermediate inference steps. A noticeable progression is evident in the model's predictions as the intermediate inference steps approach 1. The predicted actions become sharper and more aligned with the ground truth over time. Beginning from the 991-th step, the model discerns activities related to \textit{egg cooking} and \textit{juice preparation}. As the reverse diffusion progresses, the initially vague future sharpens. Consequently, the model, leveraging the visual observations, gains a deeper understanding and starts predicting actions pertinent to \textit{tea-making} with increased accuracy.
Additional qualitative results can be found in the supplementary material.
\section{Conclusion}
In this work, we have delved into the inherent challenges of predicting future actions, emphasizing the uncertainties that arise, especially when predictions extend far into the future. The proposed method, \model, leverages diffusion models to capture different possible future actions in the latent space, conditioned on the observed video. By introducing \model, we have showcased a novel approach to explicitly account for the non-deterministic nature of action anticipation. We have demonstrated the advantages of our method through extensive experiments on four benchmarks, achieving superior or comparable results to state-of-the-art methods. As the realm of action prediction continues to evolve, we believe that solutions like \model, which embrace a generative perspective, will play a pivotal role in navigating the complexities and uncertainties of future action anticipation.

{
    \small
    \bibliographystyle{ieeenat_fullname}
    \bibliography{main}
}

\clearpage
\setcounter{page}{1}
\maketitlesupplementary

\begin{figure}
    \centering
    \includegraphics[width=0.5\linewidth]{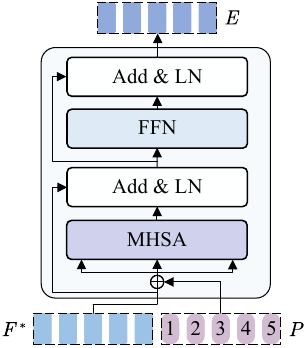}
    \caption{Details of an encoder layer.}
    \label{fig:encoder}
\end{figure}

\section{Additional Methodology Details}
\noindent\textbf{Encoder Architecture.} Our encoder follows the transformer encoder used in DETR~\cite{carion2020end} and FUTR~\cite{gongFutureTransformerLongterm2022}, as depicted in Fig.~\ref{fig:encoder}. 
The original video features $F\in \mathbb{R}^{L\times K}$ are first passed through a linear layer, which adjusts them to a suitable hidden dimension $D$, yielding transformed features $F^*\in \mathbb{R}^{L\times D}$. 
Subsequently, these transformed features are combined with sinusoidal positional encodings $P \in \mathbb{R}^{L\times D}$, facilitating the encoder in producing refined representations $E \in \mathbb{R}^{L\times D}$ via self-attention~\cite{vaswaniAttentionAllYou2017}.

\myparagraph{Training Objective.}
Our training objective consists of a past observation recognition loss $\mathcal{L}_{\text{seg}}$, a temporal smoothness loss $\mathcal{L}_{\text{smooth}}$, and a variation lower bound loss $\mathcal{L}_\text{VLB}$:
\begin{equation*}
    \mathcal{L} = \mathcal{L}_{\text{seg}} + \mathcal{L}_{\text{smooth}} + \mathcal{L}_\text{VLB}.
\end{equation*}
$\mathcal{L}_{\text{seg}}$ is a cross-entropy loss and $\mathcal{L}_{\text{smooth}}$ is computed as the mean squared error of the log-likelihoods between adjacent video frames to promote the local similarity along the temporal dimension.
We compute the variational lower bound ($\mathcal{L}_\text{VLB}$) following the original diffusion process:
\makeatletter
\renewcommand{\maketag@@@}[1]{\hbox{\m@th\normalsize\normalfont#1}}%
\makeatother
\begin{small}
\begin{align}
\mathcal{L}_\text{VLB}
&=\mathbb{E}_{ q(\mathbf{z}_{1:S}|\mathbf{z}_0)}
\Bigg[
\underbrace{\log\frac{ q(\mathbf{z}_S|\mathbf{z}_0)}{p_{\theta}(\mathbf{z}_S)}}_{\mathcal{L}^{\text{emb}}_S} + \sum_{s=2}^S \underbrace{\log{\frac{ q(\mathbf{z}_{s-1}|\mathbf{z}_0,\mathbf{z}_s)}{p_{\theta}(\mathbf{z}_{s-1}|\mathbf{z}_s,E)}}}_{\mathcal{L}^{\text{emb}}_{s-1}} \notag \\
& + {\underbrace{\log\frac{ q_{\phi}(\mathbf{z}_0|\boldsymbol{a})}{ p_{\theta}(\mathbf{z}_0|\mathbf{z}_1,E)}}_{\mathcal{L}^{\text{emb}}_0}}\underbrace{-\log p_{\theta}(\boldsymbol{a}|\mathbf{z}_0,E)\vphantom{\log{\frac{ q(\mathbf{z}_{s-1}|\mathbf{z}_0,\mathbf{z}_s)}{p_{\theta}(\mathbf{z}_{s-1}|\mathbf{z}_s)}}}}_{\mathcal{L}^\text{pred}} \Bigg].
\end{align}
\end{small}
Following~\cite{li2022diffusionlm,gong2023diffuseq}, we further simplify the training objective as follows:
\begin{equation}
    \mathcal{L}_{\text{VLB}} = \sum_{s=1}^S \mathcal{L}^{\text{emb}}_{s-1} 
- \log p_{\theta}(\boldsymbol{a}|\mathbf{z}_0,E),
\end{equation}
$$\hspace{-1mm}\text{where } \mathcal{L}^{\text{emb}}_{s-1} = 
\begin{cases} 
||\mathbf{z}_0 - f_{\theta}(\mathbf{z}_s, s,E)||^2 & \text{if } 2 \!\leq\! s \!\leq\! S \\
||\textsc{Emb}(\boldsymbol{a}) - f_{\theta}(\mathbf{z}_1, 1,E)||^2 & \text{if } s = 1,
\end{cases}$$ and
$- \log p_{\theta}(\boldsymbol{a}|\mathbf{z}_0,E) = - \log p_{\theta}(\boldsymbol{a}^c|\mathbf{z}_0,E) - \log p_{\theta}(\boldsymbol{a}^t|\mathbf{z}_0,E)$, under the assumption that action classes and durations are conditionally independent for mathematical convenience, as described in Section~\ref{sec:training}.

\section{Additional Implementation Details}
\noindent\textbf{Generation of the Local Attention Mask.}
Drawing from the methodology outlined in~\cite{yi2021asformer}, we conduct experiments where a hierarchical attention mask is applied to the attention matrix. This approach enables shallow layers to concentrate on neighboring frames, while deeper layers are allowed to attend to global frames. In each of the four encoder layers, we create an attention mask whose window size exponentially increases with the depth of the layer, with specific window sizes set at \{9, 33, 129, 513\}. This exponential growth is motivated by the intuition that lower layers capture finer, local details, while higher layers integrate more global, contextual information. During the self-attention computation, this mask is employed to assign minimal attention scores to positions outside the defined window, effectively minimizing their influence.

\myparagraph{Training.}
For training, we tailor the batch size and learning rate to each dataset. Specifically, 50Salads utilizes a batch size of 8 and a learning rate of 1e-3. Breakfast is set with a batch size of 64 and a learning rate of 5e-4. Both EpicKitchens and EGTEA Gaze+ share a batch size of 32, but while EpicKitchens has a learning rate of 2.5e-4, EGTEA Gaze+ uses 5e-4.  
All experiments are done with a machine equipped with 4 RTX A6000 GPUs. 

To sample diffusion step $s$ during training, we employ importance sampling~\cite{nichol2021improved} in our experiments, following~\cite{gong2023diffuseq}. In contrast to uniform sampling, the importance-weighted sampling algorithm allocates more steps to diffusion steps with a larger loss value $\mathcal{L}_s$, and fewer to others:
\begin{equation}
    \mathcal{L} = \mathbb{E}_{s\sim p_s}\left[\frac{\mathcal{L}_s}{p_s}\right], \,p_s \propto \sqrt{\mathbb{E}[\mathcal{L}_s^2]},\,\textstyle\sum_{s=1}^{S} p_s=1.
\end{equation}

\begin{table}[t]
    \centering
    \setlength\tabcolsep{2pt}
    \resizebox{\linewidth}{!}{
    \begin{tabular}{@{}lccccccccc@{}}
        \toprule
        \multirow{2}{*}{Samp.} & \multicolumn{4}{c}{$\beta$ ($\alpha$ = 0.2)} & \multicolumn{4}{c}{$\beta$ ($\alpha$ = 0.3)} \\
        \cmidrule(lr){2-5} \cmidrule(lr){6-9} 
        & 0.1 & 0.2 & 0.3 & 0.5 & 0.1 & 0.2 & 0.3 & 0.5 \\
        \midrule
         Uni. & 23.67 & 23.54 & 23.67 & 22.52 & \textbf{32.14} & \textbf{32.27} & \textbf{32.82} & \textbf{30.83} \\
         Imp. & \textbf{25.33} & \textbf{24.59} & \textbf{24.39} & \textbf{22.74} & 32.13 & 31.83 & 31.18 & 30.77 \\
         \bottomrule
    \end{tabular}}
    \caption{Comparison between uniform sampling and importance sampling on Breakfast using MoC. }
    \label{tab:step_sampling}
\end{table}

\section{Additional Analysis}
\noindent\textbf{Diffusion Step Sampling Method.} 
In Table~\ref{tab:step_sampling}, we compare importance sampling with uniform sampling on the Breakfast dataset. The advantages of importance sampling over uniform sampling are not definitively clear. While it demonstrates superior performance in scenarios with shorter observation periods, its efficacy diminishes in cases involving longer observation durations.

\begin{table}[t]
    \centering
    \setlength\tabcolsep{4pt}
    \resizebox{\linewidth}{!}{
    \begin{tabular}{@{}llccc@{}}
        \toprule
        Encoder & F1\@\{10/25/50\} & Edit & Acc. & \textcolor{gray}{MoC} \\ 
         \midrule
        DETR~\cite{carion2020end} & 24.84/21.79/16.56 & 35.19 & \textbf{54.33} & \textcolor{gray}{\textbf{31.48}}\\ 
        DETR~\cite{carion2020end} (local) & \textbf{29.53}/\textbf{25.70}/\textbf{19.14} & \textbf{36.31} & 52.50 & \textcolor{gray}{30.28} \\
        RoFormer~\cite{su2021roformer} & 7.77/5.57/3.26 & 14.26 & 37.88 & \textcolor{gray}{21.96} \\
        \bottomrule
    \end{tabular}}
    \caption{Observation segmentation performance on Breakfast with $\alpha=0.3$. For direct comparison, we also include the anticipation performance (marked as gray), averaged over all prediction horizons ($\beta$).}
    \label{tab:segmentation}
\end{table}

\myparagraph{Observation Segmentation Performance.} Our model, enhanced with a classification head attached to the encoder, also exhibits the capability to segment past observations. In line with previous work in action segmentation~\cite{farha2019ms-tcn,yi2021asformer,liu2023diffact}, we report frame-wise accuracy (Acc), edit score (Edit), and F1 scores at overlap thresholds 10\%, 25\%, 50\% (F1@{10, 25, 50}) in Table~\ref{tab:segmentation}. Accuracy assesses the results at the frame level, while the edit score and F1 scores measure the performance at the segment level. We note that comparing the segmentation performance of our model, optimized for action anticipation, to state-of-the-art action segmentation methods is not straightforward due to our model's focus on partial video inputs. 
For direct comparison, we also include the anticipation performance averaged across all prediction horizons. As indicated in Table~\ref{tab:segmentation}, variants of DETR~\cite{carion2020end} significantly surpass RoFormer~\cite{su2021roformer} across all metrics, highlightling the significance of precise past observation recognition in action anticipation. Although local attention contributes positively to observation recognition, its benefit for action anticipation is not conclusively established.

\begin{table}[t]
    \centering
    \setlength\tabcolsep{3pt}
    \resizebox{\linewidth}{!}{
    \begin{tabular}{@{}cccccc@{}}
    \toprule
       \# Steps & MoC & \# params & GPU Mem. & FLOPs & Inference time \\
       \midrule
       10 & 27.41 & 80.0M & 28.1G & 5.94G & 39.04ms\\
       25  & 27.32 & 80.0M & 28.1G & 10.09G & 94.31ms \\
       50 & 27.83 & 80.0M & 28.1G & 17.02G & 185.41ms\\
       100 & 27.87 & 80.0M & 28.1G & 30.87G & 370.78ms\\
       \bottomrule
    \end{tabular}}
    \caption{The mean over classes (MoC) averaged across all observation ratios ($\alpha=$ 0.2 and 0.3) and all prediction horizons on Breakfast, the number of parameters, the FLOPs at inference for a video of 1000 frames, the GPU memory cost during training for a batch size of 64, and the inference time on an A6000 GPU of \model.}
    \label{tab:computation}
\end{table}

\myparagraph{Computational Cost.}
Table~\ref{tab:computation} presents a comprehensive computation analysis of our model on Breakfast, capturing the mean over classes (MoC) across various prediction horizons. The model comprises 80 million parameters and utilizes 28.1 GB of GPU memory during training for a batch size of 64. We compute the floating-point operations per second (FLOPs) of \model using THOP\footnote{\url{https://github.com/Lyken17/pytorch-OpCounter}}. The FLOPs and inference time are calculated when inferring a video sequence of 1000 frames on a single RTX A6000 GPU, both of which scale with the number of diffusion inference steps.
While our model achieves optimal performance with 100 steps, maintaining an acceptable inference time for long-term action anticipation, it also produces satisfactory outcomes using just 10 steps, with a notably quick inference time of 39.04 milliseconds.

\section{Additional Ablation Results}
\begin{table*}[t]
    \centering
    \begin{subtable}[t]{\linewidth}
        \centering
        \begin{tabular}{lccccccccc}
        \toprule
        \multirow{2}{*}{Encoder} & \multicolumn{4}{c}{Breakfast $\beta$ ($\alpha$ = 0.2)} & \multicolumn{4}{c}{Breakfast $\beta$ ($\alpha$ = 0.3)} \\
        \cmidrule(lr){2-5} \cmidrule(lr){6-9}
        & 0.1 & 0.2 & 0.3 & 0.5 & 0.1 & 0.2 & 0.3 & 0.5 \\
        \midrule
        \rowcolor{Gray} DETR Encoder~\cite{carion2020end} &  \textbf{25.33} & \textbf{24.59} & \textbf{24.39} & 22.74 & \textbf{32.13} & \textbf{31.83} & \textbf{31.18} & \textbf{30.77} \\
         DETR Encoder~\cite{carion2020end} (local) & 23.39 & \textbf{24.59} & 24.28 & \textbf{23.48} & 31.01 & 30.58 & 30.38 & 29.16 \\
         RoFormer~\cite{su2021roformer} & 18.38 & 18.86 & 19.35 & 19.43 & 22.26 & 21.94 & 22.42 & 21.20 \\
         \bottomrule
    \end{tabular}
    \caption{Encoder architecture.}
    \end{subtable}
    \begin{subtable}[t]{\linewidth}
        \centering
        \begin{tabular}{@{}cccccccccccccc@{}}
        \toprule
        \multicolumn{3}{c}{Loss} & \multicolumn{4}{c}{Breakfast $\beta$ ($\alpha$ = 0.2)} & \multicolumn{4}{c}{Breakfast $\beta$ ($\alpha$ = 0.3)} & \multicolumn{3}{c}{EGTEA} \\
        \cmidrule(lr){1-3} \cmidrule(lr){4-7} \cmidrule(lr){8-11} \cmidrule(lr){12-14}
        $\mathcal{L}_\text{seg}$ & $\mathcal{L}_\text{smooth}$ & $\mathcal{L}_\text{VLB}$ & 0.1 & 0.2 & 0.3 & 0.5 & 0.1 & 0.2 & 0.3 & 0.5 & All & Freq & Rare \\
        \midrule
         \xmark & \xmark & \checkmark & 13.31 & 12.71 & 12.95 & 12.71 & 19.19 & 18.31 & 17.70 & 16.14 & 74.11 & 80.59 & 57.26\\
         \checkmark & \xmark & \checkmark & \textbf{25.35} & \textbf{25.81} & \textbf{24.83} & \textbf{23.64} & 31.66 & 31.44 & \textbf{32.17} & \textbf{30.91} & 77.07 & 83.27 & 60.95 \\
         \rowcolor{Gray} \checkmark & \checkmark & \checkmark & 25.33 & 24.59 & 24.39 & 22.74 & \textbf{32.13} & \textbf{31.83} & 31.18 & 30.77 & \textbf{77.33} & \textbf{83.47} & \textbf{61.35} \\
         \bottomrule
    \end{tabular}
    \caption{Loss terms.}
    \end{subtable}
    \begin{subtable}[t]{\linewidth}
        \centering
        \begin{tabular}{cccccccccc}
        \toprule
        \multirow{2}{*}{\# Queries} & \multicolumn{4}{c}{Breakfast $\beta$ ($\alpha$ = 0.2)} & \multicolumn{4}{c}{Breakfast $\beta$ ($\alpha$ = 0.3)} \\
        \cmidrule(lr){2-5} \cmidrule(lr){6-9} 
        & 0.1 & 0.2 & 0.3 & 0.5 & 0.1 & 0.2 & 0.3 & 0.5 \\
        \midrule
        4 & 23.78 & 23.88 & 24.02 & 23.43 & 31.55 & 31.24 & 30.57 & \textbf{32.78} \\
         6 & 25.35 & \textbf{25.05} & \textbf{25.44} & \textbf{25.33} & 31.42 & 31.20 & \textbf{31.38} & 30.57 \\
         \rowcolor{Gray} 8 & 25.33 & 24.59 & 24.39 & 22.74 & \textbf{32.13} & \textbf{31.83} & 31.18 & 30.77  \\
         10 & 24.53 & 24.85 & 24.32 & 22.22 & 30.50 & 30.01 & 30.43 & 28.57   \\
         12 & \textbf{26.27} & 24.83 & 24.09 & 23.49 & 29.18 & 28.28 & 29.39 & 26.88 \\
         \bottomrule
    \end{tabular}
    \caption{Number of action queries.}
    \end{subtable}
    \begin{subtable}[t]{\linewidth}
        \centering
        \begin{tabular}{ccccccccccccc}
        \toprule
        \multirow{2}{*}{\# Steps} & \multicolumn{4}{c}{Breakfast $\beta$ ($\alpha$ = 0.2)} & \multicolumn{4}{c}{Breakfast $\beta$ ($\alpha$ = 0.3)} & \multicolumn{3}{c}{EGTEA} \\
        \cmidrule(lr){2-5} \cmidrule(lr){6-9} \cmidrule(lr){10-12} 
        & 0.1 & 0.2 & 0.3 & 0.5 & 0.1 & 0.2 & 0.3 & 0.5 & All & Freq & Rare \\
        \midrule
        5 & 25.29 & 24.15 & 24.09 & 22.74 & 32.05 & 31.64 & 31.35 & 29.58 & 75.18 & 80.95 & 60.15 \\
        10 & 24.72 & 23.52 & 23.33 & 22.50 & \textbf{32.41} & 31.45 & 31.52 & 29.85 & 76.75 & \textbf{83.56} & 59.06 \\
         25 & 23.76 & 23.23 & 22.71 & 21.55 & 32.32 & 31.81 & 33.05 & 30.14 & 76.69 & 82.58 & 60.78 \\
         50 & 24.29 & 23.97 & 23.94 & \textbf{23.88} & 31.90 & 31.46 & \textbf{33.14} & 30.09 & 76.33 & 82.69 & 59.79 \\
         \rowcolor{Gray} 100 & \textbf{25.33} & \textbf{24.59} & \textbf{24.39} & 22.74 & 32.13 & \textbf{31.83} & 31.18 & \textbf{30.77} & \textbf{77.33} & 83.47 & \textbf{61.35}\\
         200 & 23.95 & 23.80 & 23.49 & 22.28 & 31.41 & 31.02 & 31.38 & 30.11 & 76.47 & 82.78 & 60.04 \\
         \bottomrule
    \end{tabular}
    \caption{Number of inference steps.}
    \end{subtable}
    \caption{Comprehensive results of the ablation study (Section~\ref{sec:ablation}).}
    \label{tab:ablation_study_full}
\end{table*}
In Table~\ref{tab:ablation_study_full}, we provide the comprehensive results for the ablation study (Section~\ref{sec:ablation}) with two observation ratios $\alpha \in \{0.2, 0.3\}$ for Breakfast and all sets for EGTEA Gaze+. 
The experimental trends generally show consistency across both observation ratios when examining variations in the encoder architecture and inference steps. However, there are minor differences in the effects of varying the loss terms and the number of action queries. Specifically, the temporal smoothness loss positively impacts performance across all sets on EGTEA Gaze+, but its advantage for the Breakfast dataset is less pronounced. In fact, this loss term slightly deteriorates performance in scenarios with shorter observations. Regarding the number of action queries, using six queries performs slightly better for shorter observations, whereas eight queries performs slightly better for longer observations.

\section{Additional Qualitative Results}
\begin{figure*}[t]
    \begin{minipage}{0.2\linewidth}
        Observation
    \end{minipage}
    \begin{minipage}{0.6\linewidth}
        \centering
        Prediction
    \end{minipage}
    \centering
    \begin{subfigure}[b]{\linewidth}
        \includegraphics[width=\linewidth]{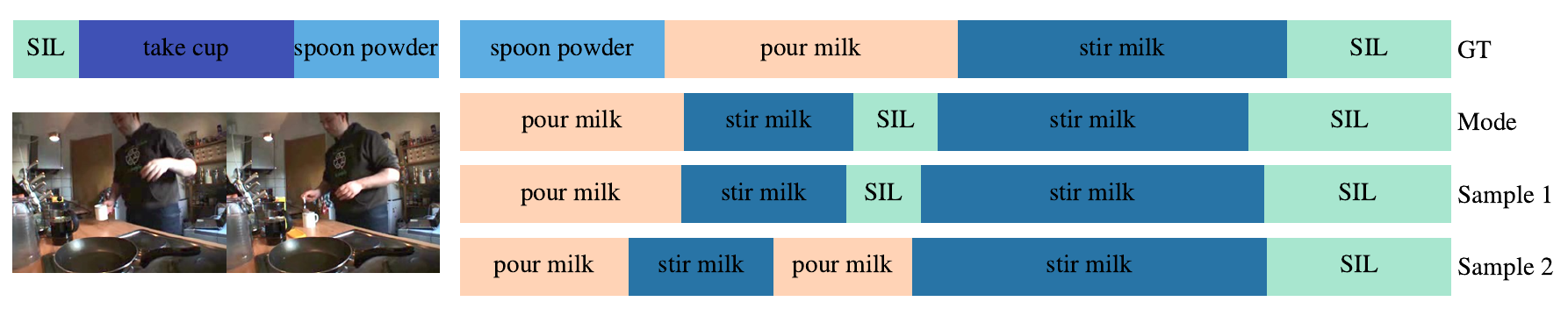}
        \subcaption{Activity: Milk}
    \end{subfigure}
    \begin{subfigure}[b]{\linewidth}
        \includegraphics[width=\linewidth]{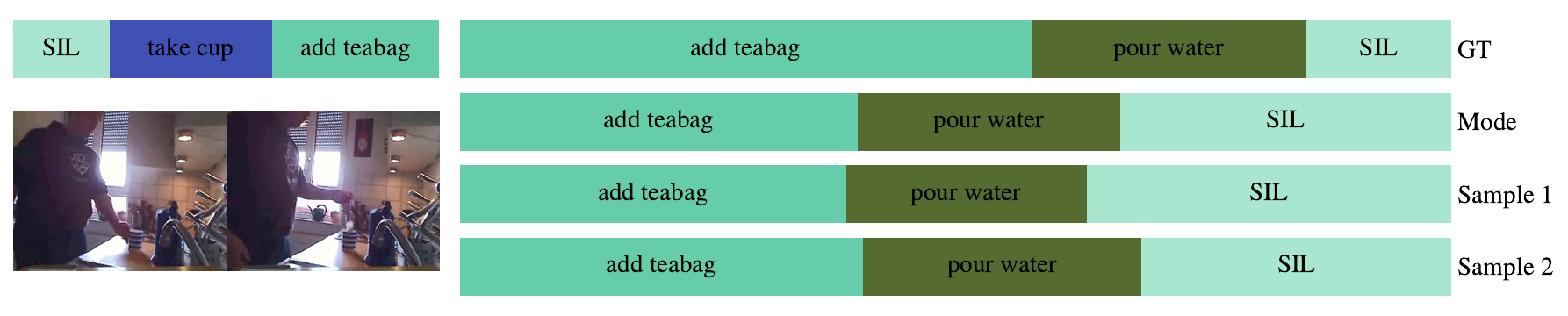}
        \subcaption{Activity: Tea}
    \end{subfigure}
    \begin{subfigure}[b]{\linewidth}
        \includegraphics[width=\linewidth]{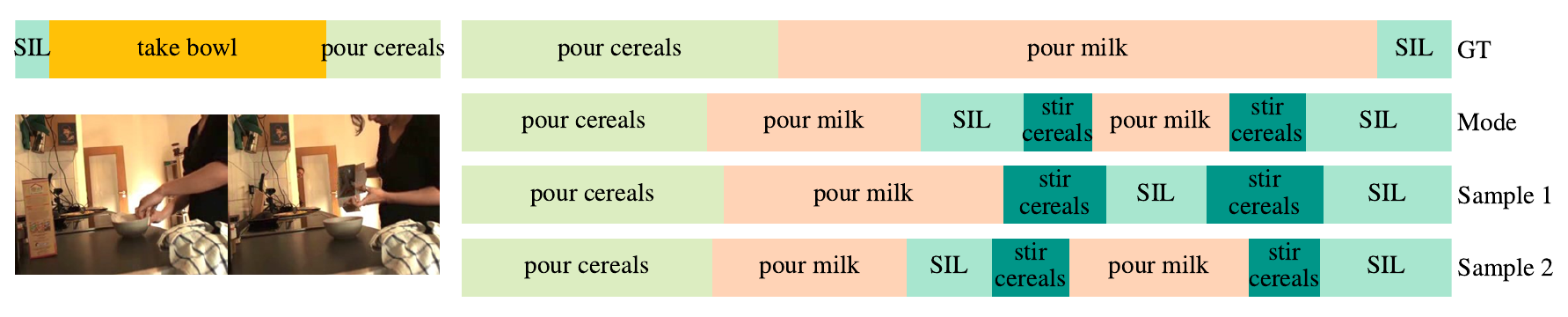}
        \subcaption{Activity: Cereals}
    \end{subfigure}
    \begin{subfigure}[b]{\linewidth}
        \includegraphics[width=\linewidth]{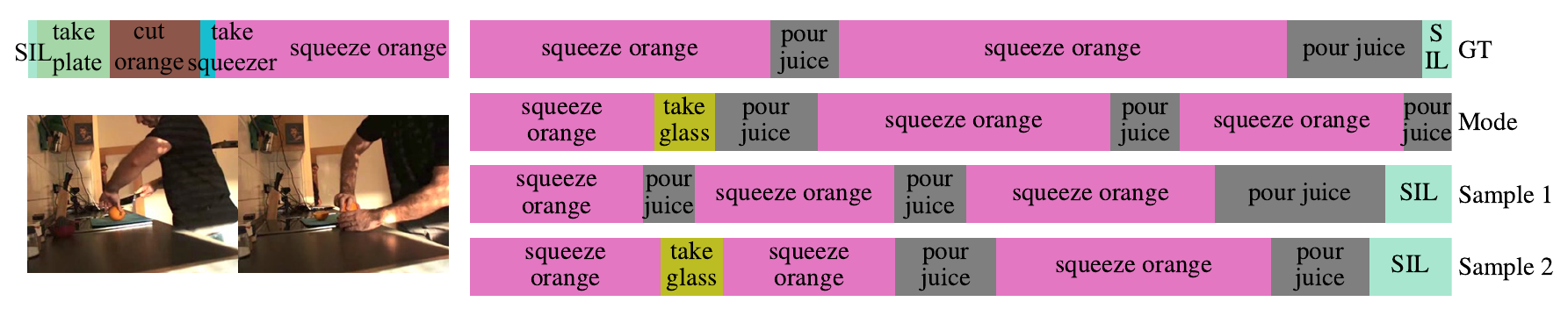}
        \subcaption{Activity: Juice}
    \end{subfigure}
    \begin{subfigure}[b]{\linewidth}
        \includegraphics[width=\linewidth]{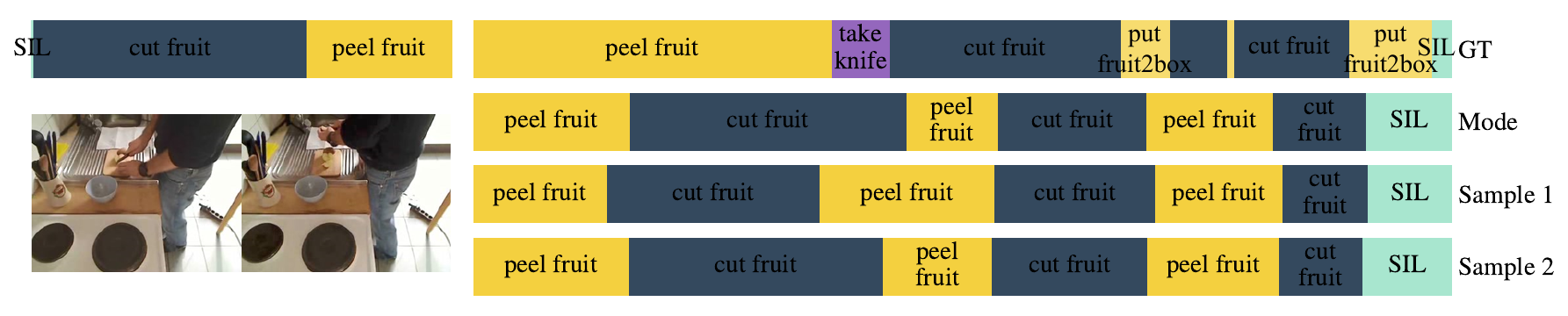}
        \subcaption{Activity: Salat}
    \end{subfigure}
    
    \caption{Additional qualitative results on Breakfast. Observations are shown on the left, while the ground-truth labels and predicted results, \ie, one deterministic prediction and two randomly sampled results, are displayed on the right. We set $\alpha$ as 0.3 and predict all subsequent actions in this experiment. Action labels and durations are decoded as frame-wise action classes.}
    \label{fig:qualitative_supp}
\end{figure*}

\begin{figure*}
    \begin{minipage}{0.2\linewidth}
        Observation
    \end{minipage}
    \begin{minipage}{0.6\linewidth}
        \centering
        Prediction
    \end{minipage}
    \centering
    \includegraphics[width=\linewidth]{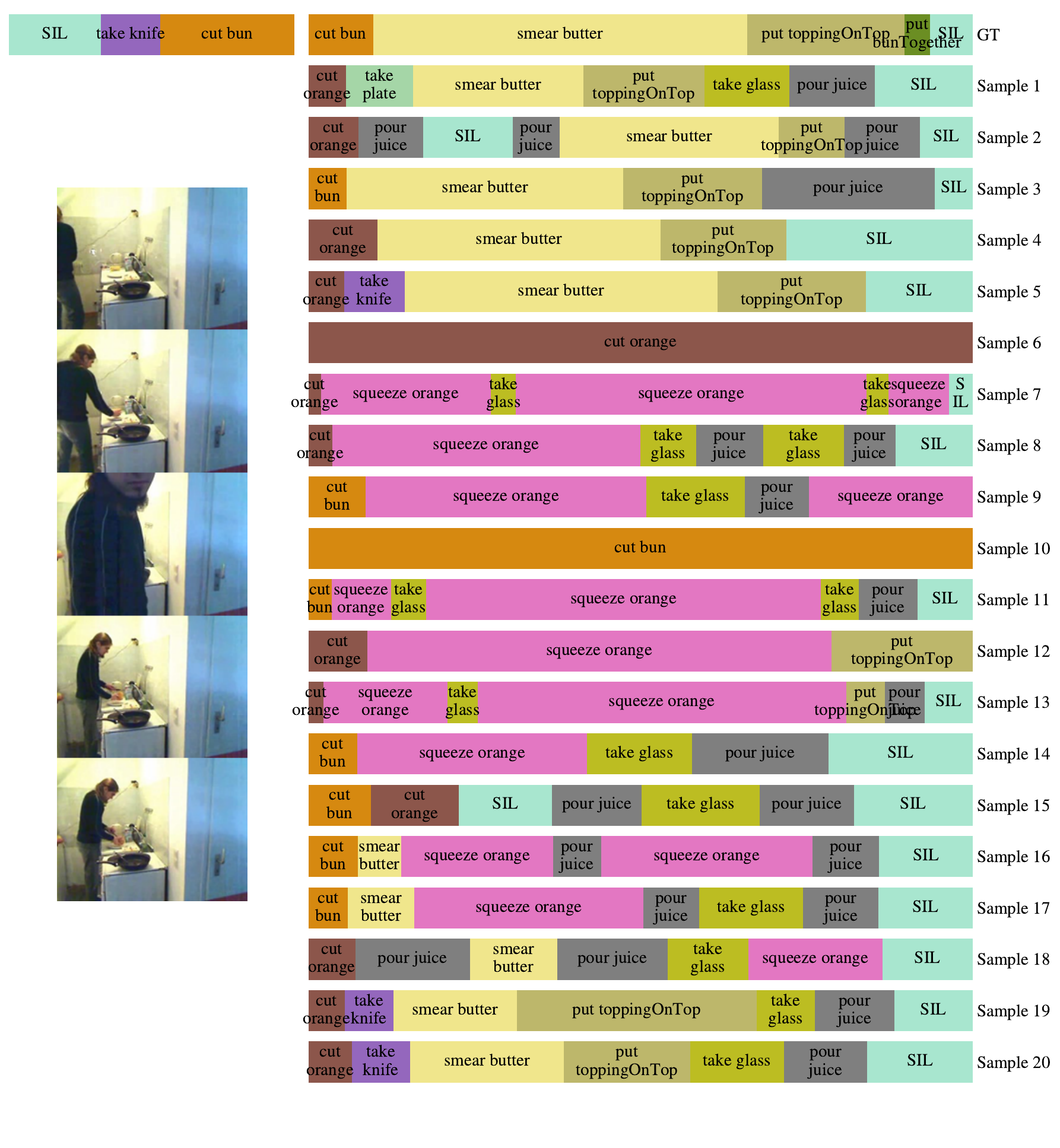}\vspace{-.8cm}
    \caption{Uncertainty-aware anticipations on Breakfast of the activity \textit{make sandwich}. Observation rate $\alpha$ is set as 0.3.}
    \label{fig:sample20_alpha0.3}
\end{figure*}

\begin{figure*}
    \begin{minipage}{0.3\linewidth}
        Observation
    \end{minipage}
    \begin{minipage}{0.6\linewidth}
        \centering
        Prediction
    \end{minipage}
    \centering
    \includegraphics[width=\linewidth]{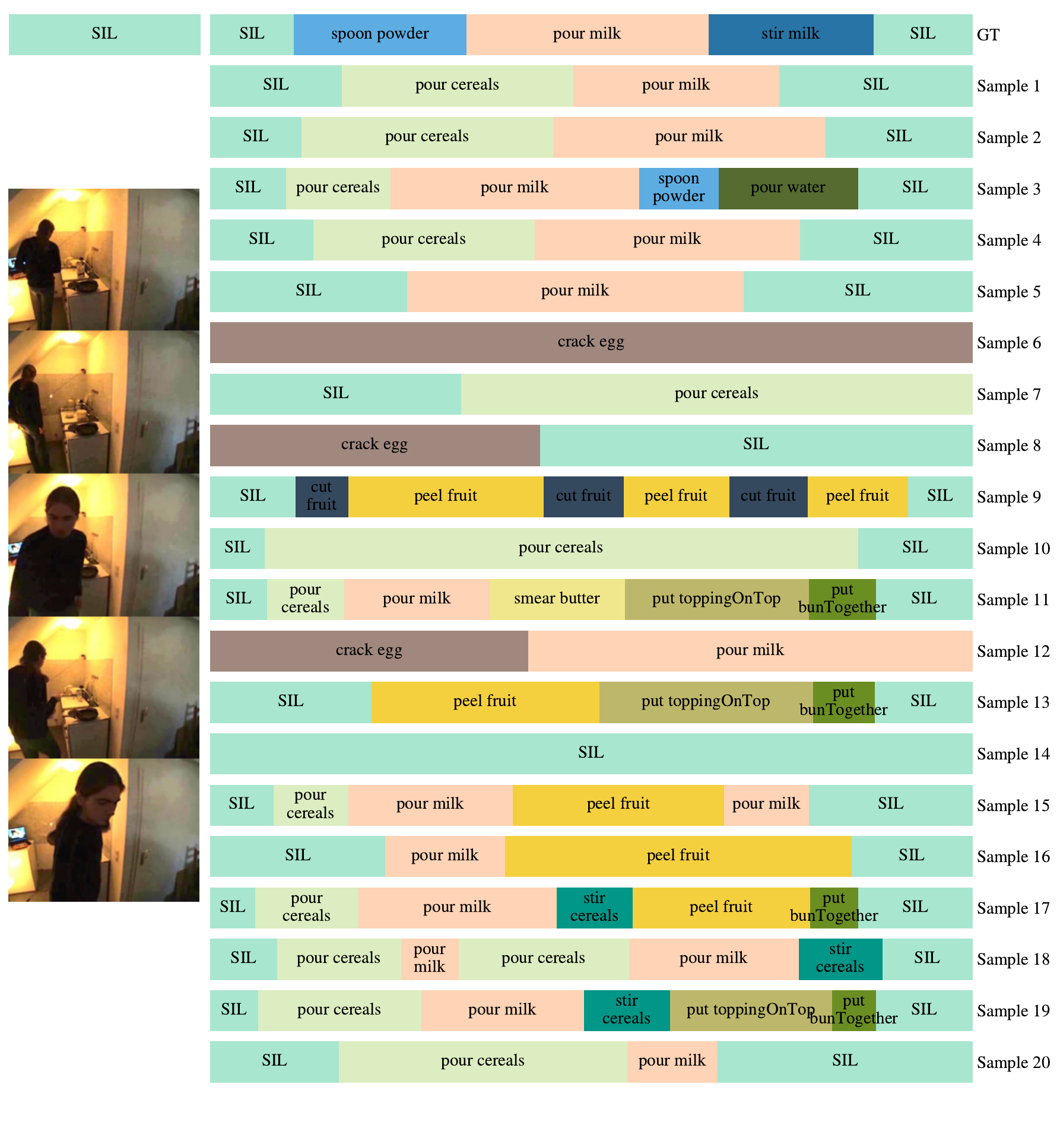}\vspace{-.8cm}
    \caption{Uncertainty-aware anticipations on Breakfast of the activity \textit{prepare milk}. Observation rate $\alpha$ is set as 0.2.}
    \label{fig:sample20_alpha0.2}
\end{figure*}

Additional qualitative results are showcased in Fig.~\ref{fig:qualitative_supp}, where the observations are displayed on the left, and the right side features the ground truth labels alongside three predictions from our model. The prediction corresponding to future action embeddings $\mathbf{z}_S$ initialized with zero vectors is denoted as \textit{Mode}. In contrast, predictions derived from randomly sampled $\mathbf{z}_S$ are identified as \textit{Sample 1} and \textit{Sample 2}. Moreover, Fig.~\ref{fig:sample20_alpha0.3} presents a collection of 20 random sample results. Given an observed video with the actions \textit{taking knife} and \textit{cutting bun}, our model successfully predicts actions associated with \textit{preparing juice} and \textit{making sandwich}, demonstrating its proficiency in generating varied yet pertinent future actions. In instances where the observation offers only limited insight into the current activity, as illustrated in Fig.~\ref{fig:sample20_alpha0.2}, our model exhibits a broader range of potential futures. These include actions such as \textit{preparing cereals}, \textit{cooking egg}, \textit{preparing fruit}, and \textit{making sandwich}, alongside the actual ground truth activity of \textit{preparing milk}.

\begin{figure*}[t]
    \begin{minipage}{0.2\linewidth}
        Observation
    \end{minipage}
    \begin{minipage}{0.6\linewidth}
        \centering
        Prediction
    \end{minipage}
    \centering
    \begin{subfigure}[b]{\linewidth}
        \includegraphics[width=\linewidth]{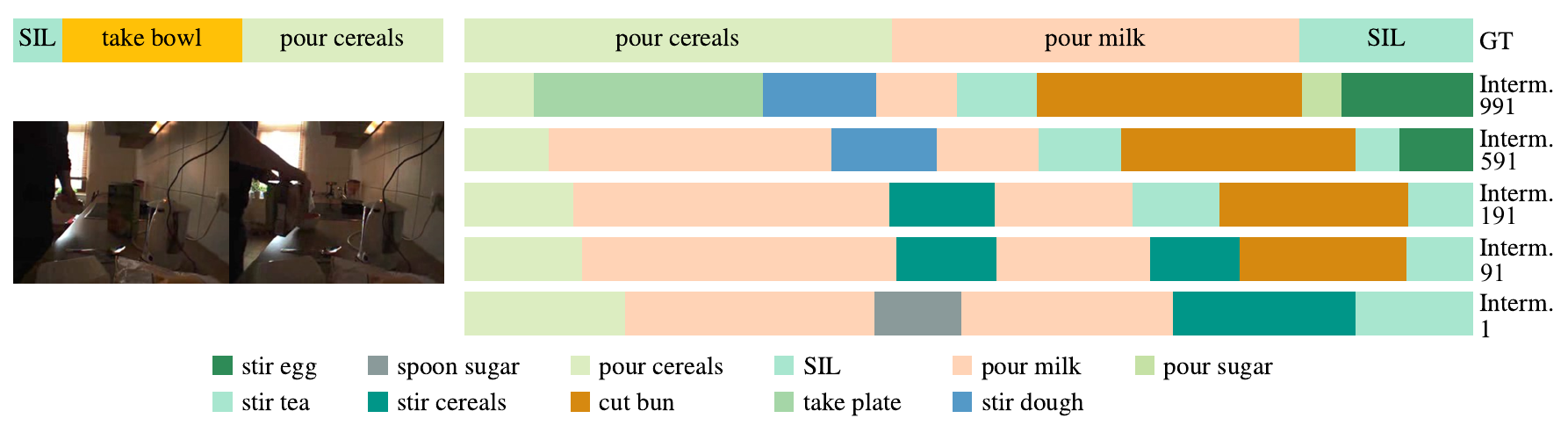}\vspace{-0.2cm}
    \subcaption{Activity: Cereals}
    \end{subfigure}
    \begin{subfigure}[b]{\linewidth}
        \includegraphics[width=\linewidth]{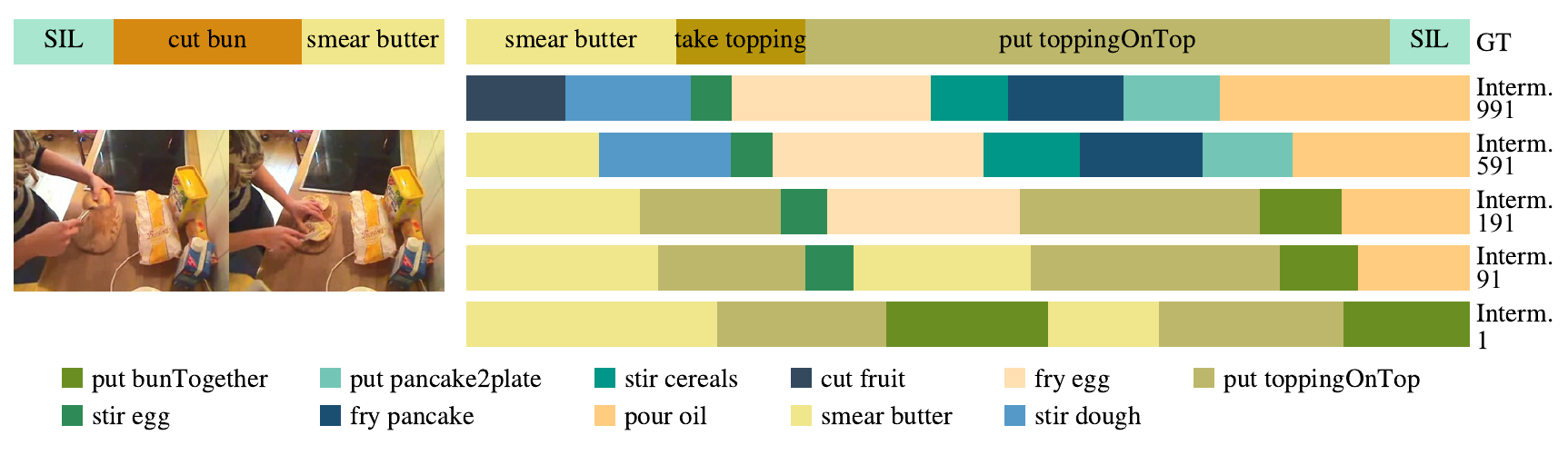}\vspace{-0.2cm}
    \subcaption{Activity: Sandwich}
    \end{subfigure}
    \begin{subfigure}[b]{\linewidth}
        \includegraphics[width=\linewidth]{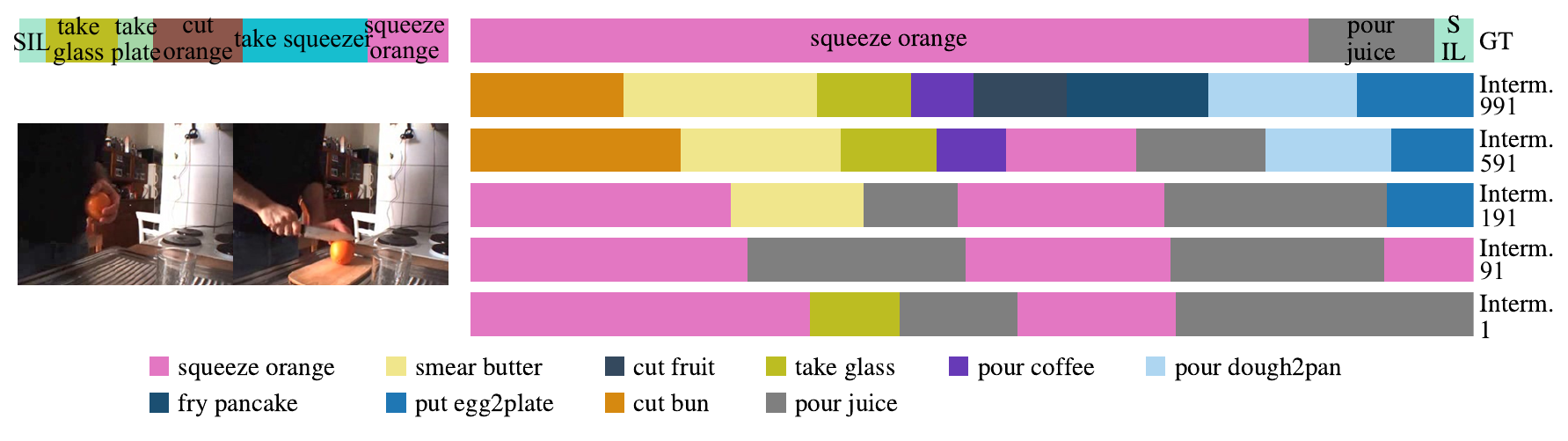}\vspace{-0.2cm}
    \subcaption{Activity: Juice}
    \end{subfigure}
    \begin{subfigure}[b]{\linewidth}
        \includegraphics[width=\linewidth]{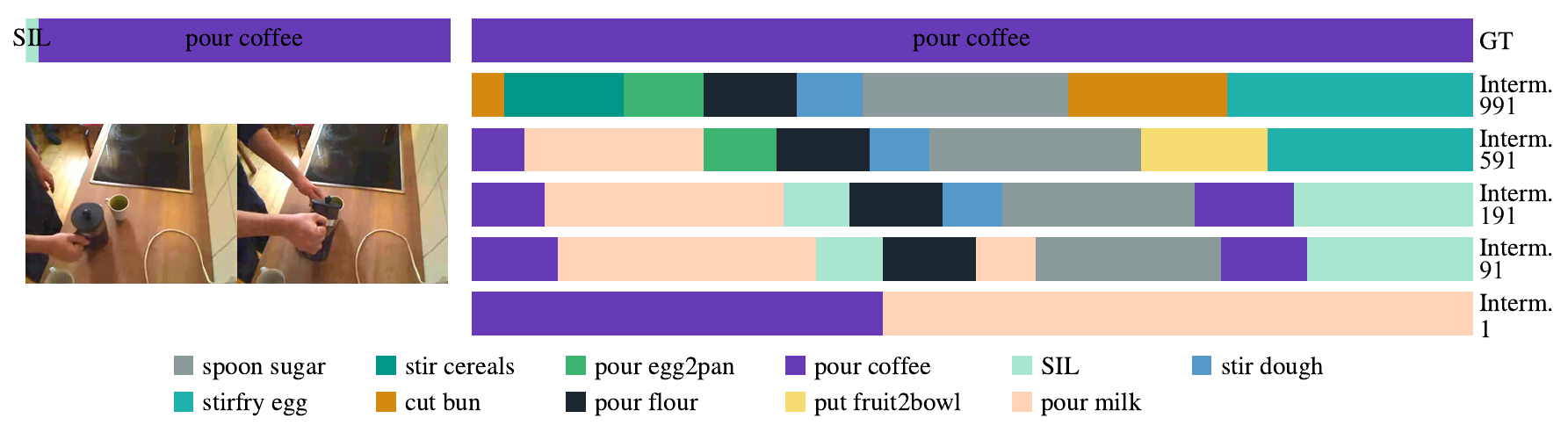}\vspace{-0.2cm}
    \subcaption{Activity: Coffee}
    \end{subfigure}\vspace{-0.2cm}
    
    \caption{Anticipation results of the intermediate inference steps on Breakfast. Observations are shown on the left, while the ground-truth labels and predicted results are displayed on the right. $\alpha$ is set as 0.3 and all subsequent actions are predicted in this experiment. As the inference step approaches 1, the predictions are gradually refined, and the ground truth actions emerge.}
    \label{fig:intermediate_supp}
\end{figure*}

Additional intermediate diffusion results are presented in Fig.~\ref{fig:intermediate_supp}. Consistent with the observations detailed in Section~\ref{sec:intermediate}, the clarity and alignment of the predicted actions with the ground truth progressively improve over time. This improvement is evident as the reverse diffusion process advances, transforming initially indistinct future predictions into more defined and accurate outcomes.

\end{document}